\documentclass[final,3p,times,twocolumn]{elsarticle}

\makeatletter
\def\ps@pprintTitle{%
  \let\@oddfoot\@empty
  \let\@evenfoot\@oddfoot
}
\makeatother

\usepackage{amssymb}
\usepackage{amsthm}

\usepackage{lineno}
\usepackage{natbib} 

\usepackage{hyperref}
\usepackage{url}

\usepackage{amsmath,amsfonts}
\usepackage{algorithmic}
\usepackage{algorithm}
\usepackage{array}
\usepackage[caption=false,font=normalsize,labelfont=sf,textfont=sf]{subfig}
\usepackage{textcomp}
\usepackage{stfloats}
\usepackage{verbatim}
\usepackage{graphicx}

\usepackage{multirow}
\usepackage{soul}
\usepackage[table]{xcolor}
\usepackage{times}
\usepackage{epsfig}
\usepackage{graphicx}
\usepackage{amssymb}
\usepackage{soul}
\usepackage{enumitem}
\usepackage{verbatim}
\usepackage{multirow}
\usepackage{float}
\usepackage[export]{adjustbox}
\usepackage{textcomp,array,booktabs}

\usepackage{wrapfig,lipsum,booktabs}

\usepackage{mathtools}
\usepackage{booktabs}

\journal{Neurocomputing}

\newcommand{\splitcell}[1]{\begin{tabular}{@{}c@{}}#1\end{tabular}}
\newcommand{\bsplitcell}[1]{$\left[\splitcell{#1}\right]$}

\newcommand{\norm}[1]{\| #1 \|}
\newcommand{\bignorm}[1]{\Bigl \| #1 \Bigr \| }

\begin{document}

\begin{frontmatter}

\title{Leveraging Angular Distributions for Improved Knowledge Distillation}

\author[a]{Eun~Som~Jeon, Hongjun~Choi,
        Ankita~Shukla,
        and~Pavan~Turaga}
\address[a]{Geometric Media Lab, School of Arts, Media and Engineering and School of Electrical, Computer and Energy Engineering, Arizona State University, Tempe, AZ, 85281, USA}

\begin{abstract}
Knowledge distillation as a broad class of methods has led to the development of lightweight and memory efficient models, using a pre-trained model with a large capacity (teacher network) to train a smaller model (student network). 
Recently, additional variations for knowledge distillation, utilizing activation maps of intermediate layers as the source of knowledge, have been studied. Generally, in computer vision applications, it is seen that the feature activation learned by a higher-capacity model contains richer knowledge, highlighting complete objects while focusing less on the background.
Based on this observation, we leverage the teacher’s dual ability to accurately distinguish between positive (relevant to the target object) and negative (irrelevant) areas.

We propose a new loss function for distillation, called angular margin-based distillation (AMD) loss. AMD loss uses the angular distance between positive and negative features by projecting them onto a hypersphere, motivated by the near angular distributions seen in many feature extractors. Then, we create a more attentive feature that is angularly distributed on the hypersphere by introducing an angular margin to the positive feature. Transferring such knowledge from the teacher network enables the student model to harness the teacher’s higher discrimination of positive and negative features, thus distilling superior student models. The proposed method is evaluated for various student-teacher network pairs on four public datasets. Furthermore, we show that the proposed method has advantages in compatibility with other learning techniques, such as using fine-grained features, augmentation, and other distillation methods. 
\end{abstract}

\begin{keyword}
Knowledge distillation, Angular distribution, Angular margin, Image classification


\end{keyword}

\end{frontmatter}



\section{Introduction}
In the past decade, convolutional neural networks (CNN) have been widely deployed into many commercial applications. Various architectures that go beyond convolutional methods have also been developed. However, a core challenge in all of them is that they are accompanied by high computational complexity, and large storage requirements \cite{gou2021knowledge, cho2019efficacy}. For this reason, application of deep networks is still limited to environments that have massive computational support. In emerging applications, there is growing demand for applying deep nets on edge, mobile, and IoT devices \cite{li2018learning, plastiras2018edge, jang2020experimental, wu2016quantized}. To move beyond these limitations, many studies have developed a lightweight form of neural models which assure performance while `lightening' the network scale \cite{cho2019efficacy, li2018learning, plastiras2018edge, jang2020experimental, wu2016quantized, han2015deep, hinton2015distilling}.
  
Knowledge distillation (KD) is one of the promising solutions that can reduce the network size and develop an efficient network model \cite{gou2021knowledge, cho2019efficacy, yim2017gift} for various fields including wearable sensor data \cite{jeon2021role}, sound \cite{tripathi2022data, li2021mutual}, and image classification \cite{wen2021preparing, chen2021knowledge}. The concept of knowledge distillation is that the network consists of two networks, a larger one called teacher and a smaller one called student \cite{hinton2015distilling}. During training the student, the teacher transfers its knowledge to the student, using the logits from the final layer. So, the student can retain the teacher model's classification performance.

Recent insights have shown that features learnt in deep-networks often exhibit an angular distribution, usually leveraged via a hyperspherical embedding \cite{choi2020amc, liu2016large, liu2017sphereface}. Such embeddings lead to improved discriminative power, and feature separability. In terms of loss-functions, these can be implemented by using angular features that correspond to the geodesic distance on the hypersphere and incorporating a preset constant margin. In this work, we show that leveraging such spherical embeddings also improves knowledge distillation.
Firstly, to get more activated features, spatial attention maps are computed and decoupled into two parts: positive and negative maps.
Secondly, we construct a new form of knowledge by projecting the features onto the hypersphere to reflect the angular distance between them. Then, we introduce an angular margin to the positive feature to get a more attentive feature representation. Finally, during the distillation, the student tries to mimic the more separated decision regions of the teacher to improve the classification performance. Therefore, the proposed method effectively regularizes the feature representation of the student network to learn informative knowledge of the teacher network.

The contributions of this paper are:
   \begin{itemize} [topsep=0pt,itemsep=-1ex,partopsep=1ex,parsep=1ex] 
   \item We propose an angular margin based distillation loss (named as AMD) which performs knowledge distillation by transferring the angular distribution of attentive features from the teacher network to the student network.
   \item We experimentally show that the proposed method results in significant improvements with different combinations of networks and outperforms other attention-based methods across four datasets of different complexities, corroborating that the performance of a higher capacity teacher model is not necessarily better.
   \item We rigorously validate the advantages of the proposed distillation method with various aspects using visualization of activation maps, classification accuracy, and reliability diagrams. 
   \end{itemize}
   
The rest of the paper is organized as follows. In section \ref{sec:related_work} and \ref{sec:background}, we describe related work and background, respectively. In section \ref{sec:proposed_method}, we provide an overview of the proposed method. In section \ref{sec:experimental_results}, we describe our experimental results and analysis. In section \ref{sec:conclusion}, we discuss our findings and conclusions.

\section{Related Work} \label{sec:related_work}
\textbf{Knowledge distillation.} Knowledge distillation, a transfer learning method, trains a smaller model by shifting knowledge from a larger model. KD is firstly introduced by Buciluǎ \emph{et al.} \cite{bucilua2006model} and is further explored by Hinton \emph{et al.} \cite{hinton2015distilling}. The main concept of KD is using soft labels by a trained teacher network. That is, mimicking soft probabilities helps students get knowledge of teachers, which improves beyond using hard labels (training labels) alone. Cho \emph{et al.} \cite{cho2019efficacy} explore which combination of student-teacher is good to obtain the better performance. They show that using a teacher trained by early stopping the training improves the efficacy of KD.
KD can be categorized into two approaches that use the outputs of the teacher \cite{gou2021knowledge}. One is response-based KD, which uses the posterior probabilities with softmax loss. The other is feature-based KD using the intermediate features with normalization. Feature-based methods can be performed with the response-based method to complement traditional KD \cite{gou2021knowledge}.
Recently, feature-based distillation methods for KD have been studied to learn richer information from the teacher for better-mimicking and performance improvement \cite{gou2021knowledge, wen2021preparing, wang2021knowledge}. Romero \emph{et al.} \cite{romero2014fitnets} firstly introduced the use of intermediate representations in FitNets using feature-based distillation. This method enables the student to mimic the teacher’s feature maps in intermediate layers. 

\textbf{Attention transfer.} \label{sec:Attention_Transfer}
To capture the better knowledge of a teacher network, attention transfer \cite{gou2021knowledge, zagoruyko2016paying, wang2019pay, ji2021show} has been utilized, which is one of the popular methods for feature-based distillation. Zagoruyko \emph{et al.} \cite{zagoruyko2016paying} suggest activation-based attention transfer (AT), which uses a sum of squared attention mapping function computing statistics across the channel dimension. Although the depth of teacher and student is different, knowledge can be transferred by the attention mapping function, which matches the depth size as one. The activation-based spatial attention maps are used as the source of knowledge for distillation with intermediate layers, where the maps are created as: $f^{d}_{sum}$($A$) = $\sum_{j=1}^{c} |A_j|^{d} $, where $f$ is a computed attention map, $A$ is an output of a layer, $c$ is the number of channels for the output, $j$ is the number for the channel, and $d$ $>$ 1. A higher value of $d$ corresponds to a heavier weight on the most discriminative parts defined by activation level.
AT (feature-based distillation method) shows better effectiveness when used with traditional KD (response-based KD) \cite{zagoruyko2016paying}. The method encourages the student to generate similar normalized maps as the teacher. However, these studies have only focused on mimicking the teacher's activation from a layer \cite{wang2021knowledge}, not considering the teacher's dual ability to accurately distinguish between positive (relevant to the target object) and negative (irrelevant). Teacher not only can generate and transfer its knowledge as an activation map directly, but also can transfer separability to distinguish between positive and negative features. We refer to this as a dual ability, which we consider for improved distillation.
The emphasized positive feature regions that encapsulate regions of the target object are crucial to predicting the correct class. In general, a higher-capacity model shows better performance, producing those regions with more attention and precision compared to the smaller network. This suggests that the transfer of distinct regions of the positive and negative pairs from teacher to student could significantly improve performance. This motivates us to focus on utilizing positive and negative pairs for extracting more attentive features, implying better separability, for distillation.

\textbf{Spherical feature embeddings.}
The majority of existing methods \cite{sun2014deep, wen2016discriminative} rely on Euclidean distance for feature distinction. These approaches could not solve the problem that classification under open-set protocol shows a meaningful result only when successfully narrowing maximal intra-class distance. To solve this problem, an angular-softmax (A-softmax) function is proposed to distinguish the features by increasing the angular margins between features \cite{liu2017sphereface}. According to its geometric interpretation, using A-softmax function equivalents to the projection of features onto the hypersphere manifold, which intrinsically matches the preliminary condition that features also lie on a manifold. Applying the angular margin penalty corresponds to the geodesic distance margin penalty in the hypersphere \cite{liu2017sphereface}. A-softmax function encourages learned features to be discriminative on hypersphere manifold. For this reason, the A-softmax function shows superior performance to the original softmax function when tested on several classification problems \cite{liu2017sphereface}. On the other hand, Choi \emph{et al.} \cite{choi2020amc} introduced angular margin based contrastive loss (AMC-loss) as an auxiliary loss, employing the discriminative angular distance metric that corresponds to geodesic distance on a hypersphere manifold. AMC-loss increases inter-class separability and intra-class compactness, improving performance in classification. The method can be combined with other deep techniques, because it easily encodes the angular distributions obtained from many types of deep feature learners \cite{choi2020amc}.

The previous methods work with logits only or work with an auxiliary loss, such as a contrastive loss. We focus on features modeled as coming from angular distributions, and focus on their separability. The observations give us an insight that the high quality features for knowledge distillation can be obtained by projecting the feature pairs onto a hypersphere. For better distillation, we construct a derive new type of implicit knowledge with positive and negative pairs from intermediate layers. The details are explained in section \ref{sec:proposed_method}.

\section{Background} \label{sec:background}

\subsection{Traditional knowledge distillation}
In standard knowledge distillation \cite{hinton2015distilling}, the loss for training a student is:
\begin{equation}
    \mathcal{L} = (1- \lambda)\mathcal{L_C} + \lambda \mathcal{L_K},
\end{equation}
where, $\mathcal{L_C}$ denotes the standard cross entropy loss, $\mathcal{L_K}$ is KD loss, and $\lambda$ is a hyperparameter; $0 < \lambda < 1$.
The error between the output of the softmax layer of a student network and the ground-truth label is penalized by the cross-entropy loss:
\begin{equation}
    \mathcal{L_{C}} = \mathcal{H}(softmax(a_{S}), y),
\end{equation}
where $\mathcal{H(\cdot)}$ is a cross entropy loss function, $a_S$ is the logits of a student (inputs to the final softmax), and $y$ is a ground truth label. 
The outputs of student and teacher are matched by KL-divergence loss:
\begin{equation}\label{eq3}
    \mathcal{L_{K}} = \tau^{2}KL(z_{T}, z_{S}),
\end{equation}
where, $z_T = softmax(a_T/\tau)$ is a softened output of a teacher network, $z_S = softmax(a_S/\tau)$ is a softened output of a student, and $\tau$ is a hyperparameter; $\tau > 1$. 
Feature distillation methods using intermediate layers can be used with the standard knowledge distillation that uses output logits. When they are used together, in general, it is beneficial to guide the student network towards inducing more similar patterns of teachers and getting a better classification performance. Thus, we also utilize the standard knowledge distillation with our proposed method. 

\subsection{Attention map}

 Denote an output as $A \in \mathcal{\mathbb{R}}^{c\times{h}\times{w}}$, where $c$ is the number of output channels, $h$ is the height for the size of output, and $w$ is width for the size of the output. The attention map for the teacher is given as follows:
 \begin{align}
     f^{l}_{T} =  \sum_{j=1}^{c} |A^{l}_{T,j}|^{2}.
 \end{align}
 Here,  $A_{T}$ is an output of a layer from a teacher model, $l$ is a specific layer, $c$ is the number of channels, $j$ is the number for the output channel, and $T$ denotes a teacher network. The attention map for the student is $f^{l'}_{S}$ =  $\sum_{j'=1}^{c'} |A^{l'}_{S,j'}|^{2} $, where $A_{S}^{l'}$ is an output of a layer from a student, $l'$ is the corresponding layer of $l$, $c'$ is the number of channels for the output, $j'$ is the number for the output channel, and $S$ denotes a student network. If the student and teacher use the same depth for transfer, $l'$ can be the layer at the same depth as $l$; if not, $l'$ can be the end of the same block for the teacher. 
 From the attention map, we obtain positive and negative maps and we project features onto hypersphere to calculate angular distance for distillation. The details are explained in section \ref{sec:proposed_method}.
 
 \subsection{Spherical feature with angular margin}
 In order to promote the learned features to have an angular distribution, \cite{liu2017sphereface, wang2018additive} proposed to introduce the angular distance between features $W$ and weights $x$. For example, $W^{T}x = \norm{W}\norm{x} cos(\theta)$, where bias is set as $0$ for simplicity, and $\theta$ is the angle between $W$ and $x$. Then, the normalization of feature and weight makes the outputs only depend on the angle between weights and features and further, $\norm{x}$ is replaced to a constant $s$ such that the features are distributed on a hypersphere with a radius of $s$. To enhance the discrimination power, angular margin $m$ is applied to the angle of the target. Finally, output logits are used to formulate probability with angular margin $m$ as below \cite{liu2017sphereface, wang2018additive}:
\begin{equation}\label{eq4}
   G^{i} = log\left (\frac
   {e^{s\cdot(cos(m\cdot\theta_{y_{i}}))}}
   {e^{s\cdot(cos(m\cdot\theta_{y_{i}}))} + \sum^{J}_{j=1,j\neq y_{i}}
   e^{s\cdot(cos(\theta_{j}))}} \right ),
\end{equation}
where, $y_{i}$ is a label and $\theta_{y_{i}}$ is a target angle for class $i$, $\theta_{j}$ is an angle obtained from $j$-th element of output logits, $s$ is a constant, and $J$ is the class number. Liu \emph{et al.} \cite{liu2017sphereface} and Wang \emph{et al.} \cite{wang2018additive} utilized output logits to obtain more discriminative features for classification on a hypersphere manifold, which performs better than using original softmax function. We use Equation \eqref{eq4}  to create the new type of feature-knowledge in the intermediate layers instead of output logits in the final classifier, thereby more attentive feature maps are transferred to the student model.

\section{Proposed Method} \label{sec:proposed_method}

\begin{figure*}[ht!]
\includegraphics[width=0.975\textwidth]{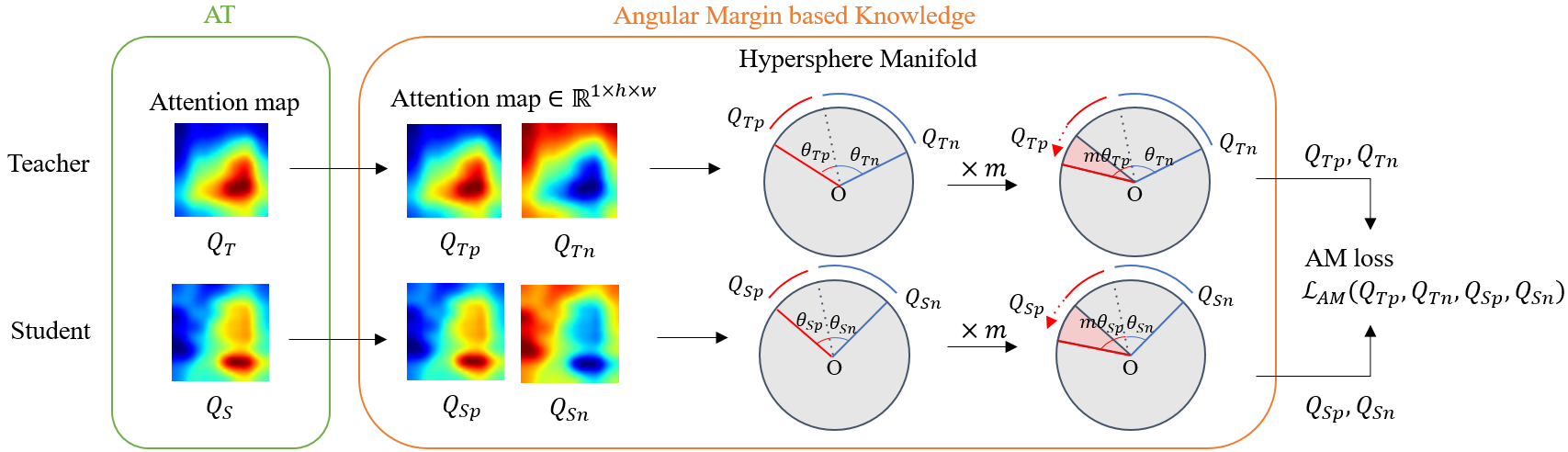}
\centering
\caption{The existing attention map-based method (AT \cite{zagoruyko2016paying}) suggested the direct use of the feature map in the intermediate layer as shown in the green box. Instead, we first decouple the feature map into the positive ($Q_p$) and negative ($Q_n$) features and map them on the hypersphere with angular margin, $m$. Then, we convert them into the probability forms and compute loss based on AM loss function. The details are explained in section \ref{proposed_AMD}.}

\label{figure:proposedAMloss}
\end{figure*}

\begin{figure}[ht!]
\includegraphics[width=0.475\textwidth]{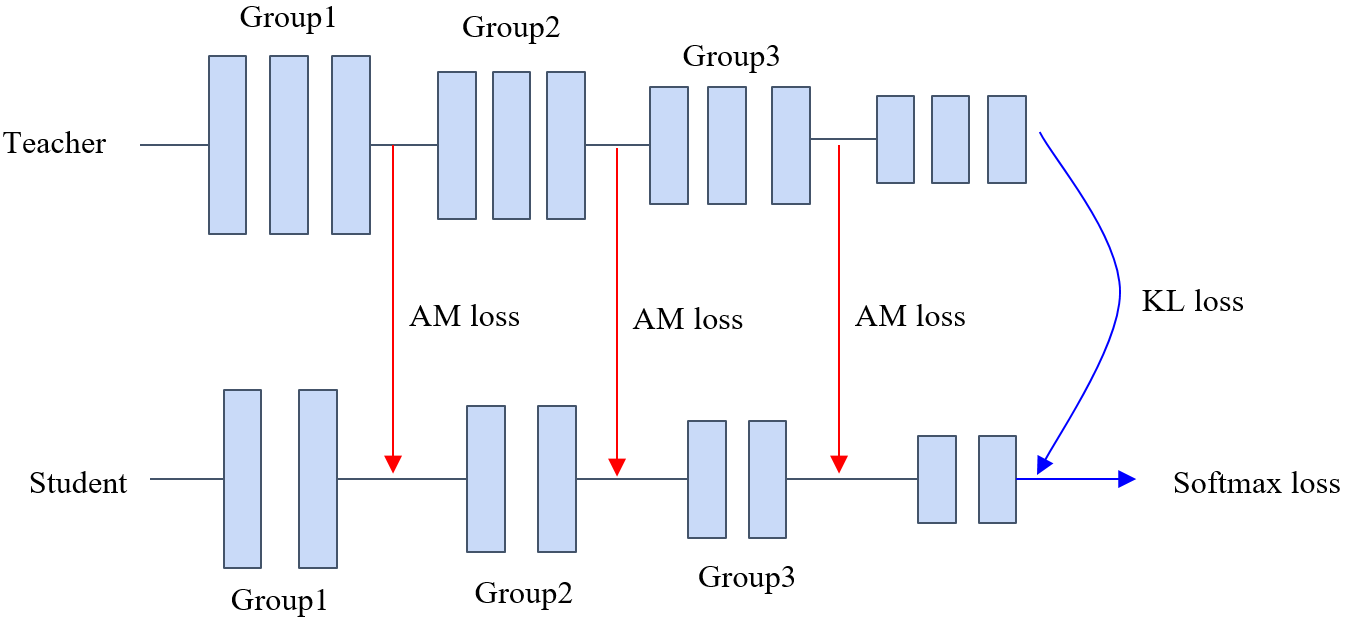}
\centering
\caption{Schematics of teacher-student knowledge transfer with the proposed method.}
\label{figure:proposedKDmethod}
\end{figure}

The proposed method utilizes features from intermediate layers of deep networks for extracting angular-margin based knowledge as illustrated in Figure \ref{figure:proposedAMloss}. The resultant angular margin loss is computed at various depths of the student and teacher as illustrated in figure \ref{figure:proposedKDmethod}. To obtain the angular distance between positive and negative features, we first generate attention maps from the outputs of intermediate layers. We then decouple the maps into positive and negative features. The features are projected onto a hypersphere to extract angularly distributed features. For effective distillation, more attentive features are obtained by introducing angular margin to the positive feature and the probability forms for distillation are computed. Finally, the knowledge of the teacher having better discrimination of positive and negative features is transferred to the student.
The details for obtaining the positive and negative maps and the angular margin based knowledge are explained in the following section.

\subsection{Generating attention maps}
To transfer activated features from teacher to student, the output of intermediate layers are used.
To match the dimension size between teacher and student models, we create the normalized attention maps \cite{zagoruyko2016paying}, which has benefits in generating maps discriminatively between positive and negative features. This reduces the need for any additional training procedure for matching the channel dimension sizes between teacher and student. We use the power value $d = 2$ for generating the attention maps, which shows the best results as reported in previous methods \cite{zagoruyko2016paying}.

\subsection{Angular margin computation} \label{proposed_AMD}
Although the activation map-based distillation provides additional context information for student model learning, there is still room to craft an attentive activation map that can distill a superior student model in KD. To further refine the original attention map, we propose an angular margin-based distillation (AMD) that encodes new knowledge using the angular distance between positive (relevant to the target object) and negative features (irrelevant) on the hypersphere.

We denote the normalized positive map as $Q_{p}=f/\norm{f}$ where $f$ is the output map extracted from the intermediate layer in networks. Further, we can obtain the normalized negative map by $Q_{n}=1-Q_{p}$. 

Then, to make the positive map more attentive, we insert an angular margin $m$ into the positive features. In this way, a new feature-knowledge encoding attentive feature can be defined as follows:
\begin{equation}
\label{eq:g}
   G^{l}(Q_{p}, Q_{n}) = 
   log\left (\frac
   {e^{s\cdot(cos(m\cdot\theta_{p_{l}}))}}
   {e^{s\cdot(cos(m\cdot\theta_{p_{l}}))} +
   e^{s\cdot(cos(\theta_{n_{l}}))}} \right ),
\end{equation} 
where, $\theta_{p_{l}}=cos^{-1}(Q_p)$ and $\theta_{n_{l}}=cos^{-1}(Q_n)$ for $l^{th}$ layer in the networks, and $m$ is a scalar angular margin. $G^{l}$ $\in \mathcal{\mathbb{R}}^{1\times{h}\times{w}}$ reflects the angular distance between positive and negative features in $l^{th}$ layer. For transferring knowledge, we aim to make the student's $G^{l}(Q_{Sp}, Q_{Sn})$ approximate the teacher's $G^{l}(Q_{Tp}, Q_{Tn})$ by minimizing the angular distance between feature maps.

\subsection{Angular margin based distillation loss}
With redesigned knowledge as above, we finally define the angular margin based distillation loss that accounts for the knowledge gap between the teacher and student activations as:
\begin{equation} \label{eq:amd}
\begin{split}
    &\mathcal{L}_{AM}(Q_{Tp},Q_{Tn}, Q_{Sp},Q_{Sn}) = \frac{1}{3|L|}
    \sum_{(l, l^\prime) \in L} \\
    &\left(  \underbracket{\bignorm{ \hat{G}^{l}(Q_{Tp}, Q_{Tn}) - \hat{G}^{l'}(Q_{Sp}, Q_{Sn})}^{2}_{F}}_\text{\normalsize\clap{\textbf{A}}} + \right.\\
     &\left. \underbracket{\bignorm{ \hat{Q}^{l}_{Tp} - \hat{Q}^{l'}_{Sp} }^{2}_{F}}_\text{\normalsize\clap{\textbf{P}}}
     + \underbracket{\bignorm{ \hat{Q}^{l}_{Tn} - \hat{Q}^{l'}_{Sn} }^{2}_{F}}_\text{\normalsize\clap{\textbf{N}}} 
     \right).
\end{split}
\end{equation}
Here, $\hat{G}$ denotes a function for normalization for output of function G, $\hat{Q}$ is a normalized map. $L$ collects the layer pairs ($l$ and $l^\prime$), and $\norm{\cdot}_F$ is the Frobenius norm \cite{tung2019similarity}. 
We will verify the performance of each component (A, P, and N) in section \ref{5_2}. 

The final loss ($\mathcal{L}_{AMD}$) of our proposed method combines all the distillation losses, including the conventional logit distillation (Equation \ref{eq3}). Thus, our overall learning objective can be written as:
\begin{equation}
    \mathcal{L}_{AMD} = \lambda_1\mathcal{L_{C}} + \lambda_2 \mathcal{L_{K}} + \gamma \mathcal{L_{A}},
\end{equation}
where, $\mathcal{L_{C}}$ is a cross-entropy loss, $\mathcal{L_{K}}$ is a knowledge distillation loss, $\mathcal{L_{A}}$ denotes the angular margin based loss from $\mathcal{L}_{AM}$, and $\lambda_1$, $\lambda_2$, and $\gamma$ are hyperparameters to control the balance between different losses.

\textbf{Global and local feature distillation.}
So far, we only consider the global feature (i.e., preserving its dimension and size). However, we point out that the global feature sometimes does not transfer more informative knowledge and rich spatial information across contexts of an input. Therefore, we also suggest utilizing local features during distillation. Specifically, the global feature is the original feature without a map division. Local features are determined by the division of the global feature. We split the global feature map from each layer by 2 for the width and height sizes of the maps to create four ($2 \times 2$) local feature maps. That is, one local map has $h/2\times w/2$ size, where $h$ and $w$ are the height and width sizes of the global map. Similar to before, local features encoding the attentive angle can be extracted for both teacher and student. Then, the losses considering global and local features for our method are:
\begin{equation}
\begin{split}
    \mathcal{L_{A_\text{global}}}=\mathcal{L}_{AM}(Q_{T},Q_{S}), \hspace{0.2cm} \\ 
    \mathcal{L_{A_\text{local}}}= \frac{1}{K} \sum^{K}_{k=1} \mathcal{L}_{AM}(Q^{k}_{T},Q^{k}_{S}),
\end{split}
\end{equation}
where $Q_{T}$ and $Q_{S}$ are global features of the teacher and student for distillation, and $Q^{k}_{T}$ and $Q^{k}_{S}$ are local features of the teacher and student, respectively, for $k$-th element of $K$, where $K$ is the total number of local maps from a map; $K$ = 4. When $\mathcal{L_{A_\text{global}}}$ and $\mathcal{L_{A_\text{local}}}$ are used together, we applied weights of 0.2 for local and 0.8 for global features to make a balance for learning.

\section{Experiments} \label{sec:experimental_results}

\begin{table}[htb!]
\centering
\caption{Description of experiments and their corresponding sections.}

\begin{center}
\scalebox{0.88}{\begin{tabular}{@{}p{7cm}|c}
\hline
\centering
Description & Section \\
\hline 
1.~Does AMD work to distill a better student? &   \multirow{5}{*}{\ref{5_2}}\\
\noindent
\vspace{-1em}
\begin{itemize}[topsep=0pt,itemsep=-1ex,partopsep=1ex,parsep=1ex] 
\item Comparison with various attention based distillation methods.
\item Investigating the effect of each component of the proposed method. \end{itemize} &  \vspace{-1em}\\ \hline
2.~What is the effect of learning with AMD from various teachers? & \multirow{3}{*}{\ref{sec:various_teacher}} \\
\noindent
\vspace{-1em}
\begin{itemize}[topsep=0pt,itemsep=-1ex,partopsep=1ex,parsep=1ex] 
\item Exploring with different capacity of teachers. \end{itemize} &  \vspace{-1em}\\ \hline
3.~What is the effect of different hyperparameters? & \multirow{2}{*}{\ref{sec:Param} }\\ 
\noindent
\vspace{-1em}
\begin{itemize}[topsep=0pt,itemsep=-1ex,partopsep=1ex,parsep=1ex] 
\item Ablation study with $\gamma$ and $m$. \vspace{-1em} \end{itemize} &  \\ \hline
4.~What are the visualized results for the area of interest? & \multirow{5}{*}{\ref{sec:activation_map}} \\ 
\noindent
\vspace{-1em}
\begin{itemize}[topsep=0pt,itemsep=-1ex,partopsep=1ex,parsep=1ex] 
\item Visualized results of activation maps from intermediate layers with or without local feature distillation. \vspace{-1em} \end{itemize} & \\ \hline
5.~Is AMD able to perform with existing methods? & \multirow{6}{*}{\ref{sec:combi_methods}} \\ 
\noindent
\vspace{-1em}
\begin{itemize}[topsep=0pt,itemsep=-1ex,partopsep=1ex,parsep=1ex] 
\item Evaluation with various methods such as fine-grained feature distillation, augmentation, and other distillation methods.
\item Generalizability analysis with ECE and reliability diagrams. \vspace{-1em} 
\end{itemize} &  \\ \hline

\end{tabular}}
\end{center}

\label{table:experiment_summary}
\end{table}


In this section, we present experimental validation of the proposed method. We evaluate the proposed method, AMD, with various combinations of teacher and student, which have different architectural styles. We run experiments on four public datasets that have different complexities. We examine the sensitivity with several different hyperparameters ($\gamma$ and $m$) for the proposed distillation and discuss which setting is the best. To demonstrate the detailed contribution, we report the results with various aspects, using classification accuracy as well as activation maps extracted by Grad-CAM \cite{selvaraju2017grad}. Finally, we investigate performance enhancement by combining previous methods including filtered feature based distillation. Each experiment and its corresponding section are described in Table \ref{table:experiment_summary}.

\subsection{Datasets}
\textbf{CIFAR-10.} CIFAR-10 dataset \cite{Krizhevsky2009} includes 10 classes with 5000 training images per class and 1000 testing images per class. Each image is an RGB image of size 32$\times$32. We use the 50000 images as the training set and 10000 as the testing set. The experiments on CIFAR-10 helps validate the efficacy of our models with less time consumption.

\textbf{CINIC-10.} We extend our experiments on CINIC-10 \cite{darlow2018cinic}. CINIC-10 comprises of augmented extension in the style of CIFAR-10, but the dataset contains 270,000 images whose scale is closer to that of ImageNet. The images are equally split into each  `train', `test', and `validate' sets. The size of the images is 32$\times$32. There are ten classes with 9000 images per class.

\textbf{Tiny-ImageNet / ImageNet.} To extend our experiments on a larger scale dataset having more complexity, we use Tiny-ImageNet \cite{le2015tiny}. The size of the images for Tiny-ImageNet is 64$\times$64. We pad them to 68$\times$68, then they are randomly cropped to 64$\times$64, and horizontally flipped, for augmentation to account for the complexity of the dataset. The training and testing sets are of size 100k and 10k respectively. The dataset includes 200 classes. 
For ImageNet \cite{deng2009imagenet}, The dataset has 1k categories with 1.2M training images. The images are randomly cropped and then resized to 224$\times$224 and horizontally flipped.

\begin{table}[]
\centering
\caption{Architecture of WRN used in experiments. Downsampling is performed in the first layers of conv3 and conv4. 16 and 28 mean depth and $k$ is width (channel multiplication) of the network.}

\begin{center}
\scalebox{0.9}{
\begin{tabular}{p{3.2em} |p{3.2em} |c |c}

\hline 
\centering
Group Name & Output Size & \multirow{2}{4.5em}{WRN16-$k$} &\multirow{2}{4.5em}{WRN28-$k$} \\
 \hline 
conv1 & 32$\times$32 & 3$\times$3, 16 & 3$\times$3, 16 \\ \hline \multirow{2}{1em}{conv2} & \multirow{2}{1em}{32$\times$32} & \multirow{2}{5.9em}{\bsplitcell{3$\times$3, 16$k$ \\ 3$\times$3, 16$k$}$\times$2} & \multirow{2}{5.9em}{\bsplitcell{3$\times$3, 16$k$ \\ 3$\times$3, 16$k$}$\times$4} \\[1.5pt]
&&&\\ \hline
\multirow{2}{1em}{conv3} & \multirow{2}{1em}{16$\times$16} & \multirow{2}{5.9em}{\bsplitcell{3$\times$3, 32$k$ \\ 3$\times$3, 32$k$}$\times$2} & \multirow{2}{5.9em}{\bsplitcell{3$\times$3, 32$k$ \\ 3$\times$3, 32$k$}$\times$4} \\ [1.5pt]
&&&\\ \hline
\multirow{2}{1em}{conv4} & \multirow{2}{1em}{8$\times$8} & \multirow{2}{5.9em}{\bsplitcell{3$\times$3, 64$k$ \\ 3$\times$3, 64$k$}$\times$2} & \multirow{2}{5.9em}{\bsplitcell{3$\times$3, 64$k$ \\ 3$\times$3, 64$k$}$\times$4} \\[1.5pt]
&&&\\ \hline
&1$\times$1&\multicolumn{2}{c}{average pool, 10-d fc, softmax}\\ \hline

\end{tabular} }
\end{center}

\label{table:WRNnetworks}
\end{table}

\subsection{Settings for experiments}

For experiments on CIFAR-10, CINIC-10, and Tiny-ImageNet, we set the batch size as 128, the total epochs as 200 using SGD with momentum 0.9, a weight decay of $1\times10^{-4}$, and the initial learning rate $lr$ as 0.1 which is decayed by a factor of 0.2 at epochs 40, 80, 120, and 160. For ImageNet, we use SGD with momentum of 0.9 and the batch size is set as 256. We run a total epoch of 100. The initial learning rate $lr$ is 0.1 decayed by 0.1 in 30, 60, and 90 epochs.

In experiments, we use the proposed method with WideResNet (WRN) \cite{zagoruyko2016wide} for teacher and student models to evaluate the classification accuracy, which is popularly used for KD \cite{cho2019efficacy, yim2017gift, zagoruyko2016paying, tung2019similarity}. Their network architectures are described in Table \ref{table:WRNnetworks}.

\begin{figure}[] 
\includegraphics[scale=0.5] {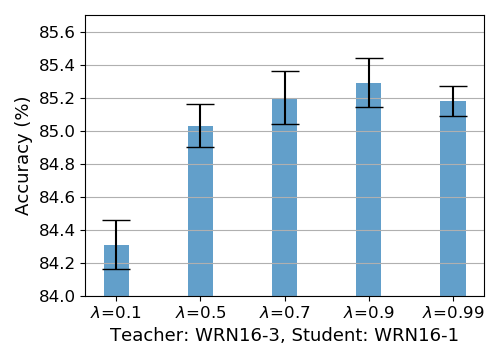}
\centering
\caption{Accuracy ($\%$) of students (WRN16-1) trained with a teacher (WRN16-3) on CIFAR-10 for various $\lambda_2$. $\lambda_1$ is obtained by 1 - $\lambda_2$.}
\label{figure:lambda_cifar10}
\end{figure}

To determine optimal parameters $\lambda_1$ and $\lambda_2$ for KD, we tested with different values for $\lambda_1$ and $\lambda_2$ for training based on KD on CIFAR-10 dataset. As shown in Figure \ref{figure:lambda_cifar10}, when $\lambda_1$ is 0.1 and $\lambda_2$ is 0.9 ($\tau$ = 4) with KD, the accuracy of a student (WRN16-1) trained with WRN16-3 as a teacher is the best. If $\lambda_1$ is small and $\lambda_2$ is large, the distillation effect of KD is increased. Since the accuracy depends on $\lambda_1$ and $\lambda_2$, we referred to previous studies \cite{cho2019efficacy, ji2021show, tung2019similarity} to choose the popular parameters for experiments. The parameters of ($\lambda_1$ = 0.1, $\lambda_2$ = 0.9, $\tau$ = 4), ($\lambda_1$ = 0.4, $\lambda_2$ = 0.6, $\tau$ = 16), ($\lambda_1$ = 0.7, $\lambda_2$ = 0.3, $\tau$ = 16), and ($\lambda_1$ = 1.0, $\lambda_2$ = 1.0, $\tau$ = 4) are used for KD on CIFAR-10, CINIC-10, Tiny-ImageNet, and ImageNet, respectively.

We perform baseline comparisons with traditional KD \cite{hinton2015distilling}, attention transfer (AT) \cite{zagoruyko2016paying}, relational knowledge distillation (RKD) \cite{park2019relational}, variational information distillation (VID) \cite{ahn2019variational}, similarity-preserving knowledge distillation (SP) \cite{tung2019similarity}, correlation congruence for knowledge distillation (CC) \cite{peng2019correlation}, contrastive representation distillation (CRD) \cite{tian2019contrastive}, attentive feature distillation and selection (AFDS) \cite{wang2019pay}, and attention-based feature distillation (AFD) \cite{ji2021show} that is a new feature linking method considering similarities between the teacher and student features, including state-of-the-art approaches. Note that, for fair comparison, the distillation methods are performed with traditional KD to see if they enhance standard KD, keeping the same setting as the proposed method. The hyperparameters of the methods follow their respective papers. For the proposed method, the constant parameter $s$ and margin parameter $m$ are 64 and 1.35, respectively. The loss weight $\gamma$ of the proposed method is 5000. We determine the hyperparameters empirically, considering the distillation effects by the capacity of models. A more detailed description of parameters appears in section \ref{sec:Param}. All experiments were repeated five times, and the averaged best accuracy and the standard deviation of performance are reported.

No augmentation method is applied for CIFAR-10 and CINIC-10. 
For the proposed method, additional techniques, such as using the other hidden layers for generating better distillation effects from teachers or reshaping the dimension size of the feature maps, are not applied. All of our experiments are run on a 3.50 GHz CPU (Intel® Xeon(R) CPU E5-1650 v3), 48 GB memory, and NVIDIA TITAN Xp (3840 NVIDIA® CUDA® cores and 12 GB memory) graphic card \cite{gpuspec}.

\begin{figure}[] 
\includegraphics[scale=0.46] {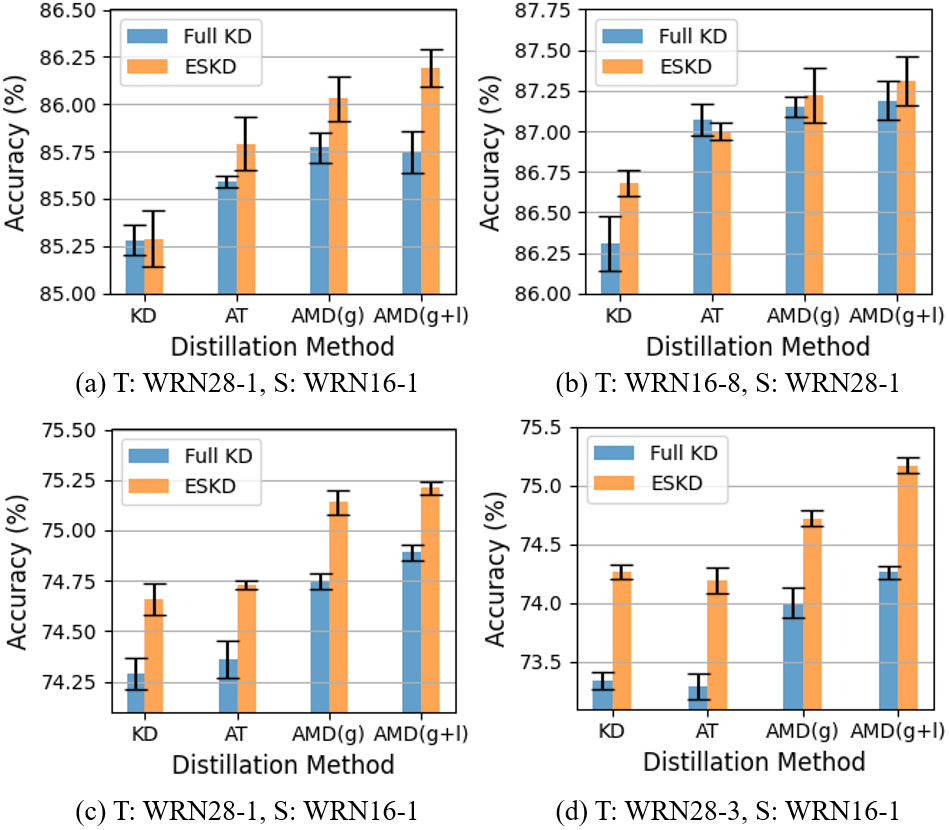}
\centering
\caption{Accuracy ($\%$) for Full KD and ESKD. (a) and (b) are on CIFAR-10, and (c) and (d) are on CINIC-10, respectively. T and S denotes teacher and student models, respectively.}
\label{figure:amd_eskd}
\end{figure}

To obtain the best performance, we adopt early-stopped KD (ESKD) \cite{cho2019efficacy} for training teacher and student models, leveraging its effects across the board in improving the efficacy of knowledge distillation. As shown in Figure \ref{figure:amd_eskd}, the early stopped model of a teacher tends to train student models better than Full KD that uses a fully trained teacher. 

\subsection{Attention-based distillation}\label{5_2}

\begin{table*}[htb!]
\centering
\caption{Details of teacher and student network architectures. ResNet \cite{he2016deep} and WideResNet \cite{zagoruyko2016wide} are denoted by ResNet (depth) and WRN (depth)-(channel multiplication), respectively.}

\begin{center}
\scalebox{0.83}{\begin{tabular}{c |c |c |c |c |c |c |c| c| c}
\hline
\centering
\multirow{2}{*}{DB}&  \multirow{2}{*}{Setup} &  \multirow{2}{*}{Compression type} & \multirow{2}{*}{Teacher} & \multirow{2}{*}{Student} & FLOPs & FLOPs &\# of params  &\# of params & Compression  \\
& & & & & (teacher) & (student) & (teacher) & (student) & ratio  \\
 \hline 
\multirow{4}{*}{\rotatebox[origin=c]{90}{\scriptsize{CIFAR-10}}}&(a) & Channel & WRN16-3 & WRN16-1 & 224.63M & 27.24M & 1.50M & 0.18M & 11.30$\%$ \\ 
&(b) & Depth & WRN28-1 & WRN16-1 & 56.07M & 27.24M & 0.37M & 0.18M & 47.38$\%$ \\ 
&(c) & Depth+Channel & WRN16-3 & WRN28-1 & 224.63M & 56.07M & 1.50M & 0.37M & 23.85$\%$ \\ 
&(d) & Different architecture & ResNet44 & WRN16-1 & 99.34M & 27.24M & 0.66M & 0.18M & 26.47$\%$ \\ 
\hline

\multirow{4}{*}{\rotatebox[origin=c]{90}{\scriptsize{CINIC-10}}}&(a) & Channel & WRN16-3 & \multirow{4}{*}{WRN16-1} & 224.63M & \multirow{4}{*}{27.24M} & 1.50M & \multirow{4}{*}{0.18M} & 11.30$\%$ \\ 
&(b) & Depth & WRN28-1 & & 56.07M &  & 0.37M & & 47.38$\%$ \\ 
&(c$^a$) & Depth+Channel & WRN28-3 & & 480.98M & & 3.29M & & 5.31$\%$ \\ 
&(d) & Different architecture & ResNet44 & & 99.34M &  & 0.66M & & 26.47$\%$ \\ 
\hline

\multirow{4}{*}{\rotatebox[origin=c]{90}{\scriptsize{Tiny-ImageNet}}}&(a) & Channel & WRN16-3 & \multirow{4}{*}{WRN16-1} & 898.55M & \multirow{4}{*}{108.98M} & 1.59M & \multirow{4}{*}{0.19M} & 11.82$\%$ \\ 
&(b$^b$) & Depth & WRN40-1 & & 339.60M &  & 0.58M & & 32.52$\%$ \\
&(c$^b$) & Depth+Channel & WRN40-2 & & 1,323.10M & & 2.27M & & 8.26$\%$ \\ 
&(d) & Different architecture & ResNet44 & & 397.36M &  & 0.67M & & 27.82$\%$ \\ 
\hline

\end{tabular}}
\end{center}

\label{table:info_settings}
\end{table*}

\begin{table*}[htb!]
\centering
\caption{Accuracy ($\%$) on CIFAR-10 with various knowledge distillation methods. The methods denoted by ``*'' are attention based distillation. ``$\mathrm{g}$'' and ``$\mathrm{l}$'' denote using global and local feature distillation, respectively.}

\begin{center}
\scalebox{0.93}{
\begin{tabular}{c |c c | c c c c c c c | c c}

\hline
\centering
\multirow{3}{*}{Setup} & \multicolumn{11}{c}{Method} \\ \cline{2-12} 
& \multirow{2}{*}{Teacher} & \multirow{2}{*}{Student} & \multirow{2}{*}{KD} & \multirow{2}{*}{AT$^*$} & \multirow{2}{*}{SP} & \multirow{2}{*}{RKD} & \multirow{2}{*}{VID} & \multirow{2}{*}{AFDS$^*$} & \multirow{2}{*}{AFD$^*$} & \multicolumn{2}{c}{AMD} \\
&&&&&&&&&&(g) & (g+l) \\ \hline

\multirow{2}{*}{(a)} & 87.76 & 84.11 & 85.29 & 85.79 & 85.69 & 85.45 & 85.40 & \multirow{2}{*}{--} & 86.23 & 86.28 & \textbf{86.36} \\
& \scriptsize$\pm$0.12 & \scriptsize$\pm$0.12 & \scriptsize$\pm$0.15 & \scriptsize$\pm$0.14 & \scriptsize$\pm$0.11 &\scriptsize$\pm$0.09 & \scriptsize$\pm$0.14 & & \scriptsize$\pm$0.13 &\scriptsize$\pm$0.06 &\scriptsize$\pm$0.10 \\

\multirow{2}{*}{(b)} & 85.59 & 84.11 & 85.48 & 85.79 & 85.77 & 85.47 & 84.92 & 85.53 & 85.84 & 86.04 & \textbf{86.10} \\
& \scriptsize$\pm$0.13 & \scriptsize$\pm$0.12 & \scriptsize$\pm$0.12 & \scriptsize$\pm$0.12 & \scriptsize$\pm$0.07 &\scriptsize$\pm$0.12 & \scriptsize$\pm$0.13 &\scriptsize$\pm$0.13  & \scriptsize$\pm$0.11 &\scriptsize$\pm$0.12 &\scriptsize$\pm$0.10 \\

\multirow{2}{*}{(c)} & 87.76 & 85.59 & 86.57 & 86.77 & 86.56 & 86.38 & 86.64 & \multirow{2}{*}{--} & 87.24 & 87.13 & \textbf{87.35} \\
& \scriptsize$\pm$0.12 & \scriptsize$\pm$0.12 & \scriptsize$\pm$0.16 & \scriptsize$\pm$0.11 & \scriptsize$\pm$0.09 &\scriptsize$\pm$0.22 & \scriptsize$\pm$0.24 & & \scriptsize$\pm$0.03 &\scriptsize$\pm$0.14 &\scriptsize$\pm$0.10 \\

\multirow{2}{*}{(d)} & 86.41 & 84.11 & 85.44 & 85.95 & 85.41 & 85.50 &85.17 & 85.14 & 85.78 & 86.22 & \textbf{86.34} \\
& \scriptsize$\pm$0.20 & \scriptsize$\pm$0.21 & \scriptsize$\pm$0.06 & \scriptsize$\pm$0.05 & \scriptsize$\pm$0.12 &\scriptsize$\pm$0.06 & \scriptsize$\pm$0.11 &\scriptsize$\pm$0.13  & \scriptsize$\pm$0.09 &\scriptsize$\pm$0.07 &\scriptsize$\pm$0.05 \\

\hline

\end{tabular} }
\end{center}

\label{table:att_CIFAR10}
\end{table*}


\begin{table*}[htb!] 
\centering
\caption{Accuracy ($\%$) on CINIC-10 with various knowledge distillation methods. The methods denoted by ``*'' are attention based distillation. AMD outperforms RKD \cite{park2019relational}. ``$\mathrm{g}$'' and ``$\mathrm{l}$'' denote using global and local feature distillation, respectively. }

\begin{center}
\scalebox{0.93}{
\begin{tabular}{c |c c | c c c c c c | c c}

\hline
\centering
\multirow{3}{*}{Setup} & \multicolumn{10}{c}{Method} \\ \cline{2-11} 
& \multirow{2}{*}{Teacher} & \multirow{2}{*}{Student} & \multirow{2}{*}{KD} & \multirow{2}{*}{AT$^*$} & \multirow{2}{*}{SP} & \multirow{2}{*}{VID} & \multirow{2}{*}{AFDS$^*$} & \multirow{2}{*}{AFD$^*$} & \multicolumn{2}{c}{AMD} \\
&&&&&&&&&(g) & (g+l) \\ \hline

\multirow{2}{*}{(a)} & 75.40 & \multirow{8}{*}{\shortstack{72.05 \\ {\scriptsize$\pm$0.12}}} & 74.31 & 74.63 & 74.43 & 74.35 & \multirow{2}{*}{--} & 74.13 & 75.04 & \textbf{75.18} \\
& \scriptsize$\pm$0.12 & & \scriptsize$\pm$0.10 & \scriptsize$\pm$0.13 & \scriptsize$\pm$0.14 &\scriptsize$\pm$0.05 & & \scriptsize$\pm$0.12 &\scriptsize$\pm$0.11 &\scriptsize$\pm$0.09 \\

\multirow{2}{*}{(b)} & 75.59 &  & 74.66 & 74.73 & 74.94 & 73.85 & 74.54 & 74.36 & 75.14 & \textbf{75.21} \\
& \scriptsize$\pm$0.15 &  & \scriptsize$\pm$0.08 & \scriptsize$\pm$0.02 & \scriptsize$\pm$0.11 &\scriptsize$\pm$0.08 & \scriptsize$\pm$0.08 & \scriptsize$\pm$0.04 &\scriptsize$\pm$0.06 &\scriptsize$\pm$0.04 \\

\multirow{2}{*}{(c$^a$)} & 76.97 &  & 74.26 & 74.19 & 75.05 & 74.06 & \multirow{2}{*}{--} & 74.20 & 74.72 & \textbf{75.17} \\
& \scriptsize$\pm$0.05 &  & \scriptsize$\pm$0.06 & \scriptsize$\pm$0.11 & \scriptsize$\pm$0.10 &\scriptsize$\pm$0.15 & & \scriptsize$\pm$0.12 &\scriptsize$\pm$0.07 &\scriptsize$\pm$0.07 \\

\multirow{2}{*}{(d)} & 74.30 &  & 74.47 & 74.67 & 74.46 & 74.43 & 74.64 & 73.31 & 74.93 & \textbf{75.10} \\
& \scriptsize$\pm$0.15 &  & \scriptsize$\pm$0.09 & \scriptsize$\pm$0.05 & \scriptsize$\pm$0.17 &\scriptsize$\pm$0.10 & \scriptsize$\pm$0.12 &\scriptsize$\pm$0.13  & \scriptsize$\pm$0.07 &\scriptsize$\pm$0.10 \\

\hline

\end{tabular} }
\end{center}

\label{table:att_CINIC10}
\end{table*}


\begin{table*}[htb!]
\centering
\caption{Accuracy ($\%$) on Tiny-ImageNet with various knowledge distillation methods. The methods denoted by ``*'' are attention based distillation. AMD outperforms VID \cite{ahn2019variational} and RKD \cite{park2019relational}. ``$\mathrm{g}$'' and ``$\mathrm{l}$'' denote using global and local feature distillation, respectively.}

\begin{center}
\scalebox{0.93}{
\begin{tabular}{c |c c | c c c c c | c c}

\hline
\centering
\multirow{3}{*}{Setup} & \multicolumn{9}{c}{Method} \\ \cline{2-10} 
& \multirow{2}{*}{Teacher} & \multirow{2}{*}{Student} & \multirow{2}{*}{KD} & \multirow{2}{*}{AT$^*$} & \multirow{2}{*}{SP} & \multirow{2}{*}{AFDS$^*$} & \multirow{2}{*}{AFD$^*$} & \multicolumn{2}{c}{AMD} \\
&&&&&&&&(g) & (g+l) \\ \hline

\multirow{2}{*}{(a)} & 58.16 & \multirow{8}{*}{\shortstack{49.45 \\ {\scriptsize$\pm$0.20}}} & 49.99 & 49.72 & 49.27 & \multirow{2}{*}{--} & 50.00 & \textbf{50.32} & 49.92 \\
& \scriptsize$\pm$0.30 & & \scriptsize$\pm$0.15 & \scriptsize$\pm$0.15 & \scriptsize$\pm$0.19 & &\scriptsize$\pm$0.23 &\scriptsize$\pm$0.07  &\scriptsize$\pm$0.04\\

\multirow{2}{*}{(b$^b$)} & 54.74 & & 49.56 & 49.79 & 49.89 & 49.46 & 50.04 & \textbf{50.15} & 49.97 \\
& \scriptsize$\pm$0.24 & & \scriptsize$\pm$0.17 & \scriptsize$\pm$0.22 & \scriptsize$\pm$0.20 &\scriptsize$\pm$0.28 & \scriptsize$\pm$0.27 & \scriptsize$\pm$0.10 &\scriptsize$\pm$0.18 \\

\multirow{2}{*}{(c$^b$)} & 59.92 & & 49.67 & 49.62 & 49.59 & \multirow{2}{*}{--} & 49.78 & 49.88 & \textbf{50.07} \\
& \scriptsize$\pm$0.15 & & \scriptsize$\pm$0.13 & \scriptsize$\pm$0.16 & \scriptsize$\pm$0.25 & &\scriptsize$\pm$0.24&\scriptsize$\pm$0.20 &\scriptsize$\pm$0.10 \\

\multirow{2}{*}{(d)} & 54.66 & & 49.52 & 49.45 & 49.13 & 49.55 & 49.44 & 49.92 & \textbf{50.08} \\
& \scriptsize$\pm$0.14 & & \scriptsize$\pm$0.16 & \scriptsize$\pm$0.28 & \scriptsize$\pm$0.20 &\scriptsize$\pm$0.13 & \scriptsize$\pm$0.27 &\scriptsize$\pm$0.09  & \scriptsize$\pm$0.16 \\

\hline

\end{tabular} }
\end{center}

\label{table:att_Tiny}
\end{table*}


\begin{table*}[htb!]
\caption{Top-1 and Top-5 accuracy (\%) on ImageNet with various knowledge distillation methods. The methods denoted by ``*'' are attention based distillation. ``$\mathrm{g}$'' and ``$\mathrm{l}$'' denote using global and local feature distillation, respectively.}
\label{test_accuracy_table_on_imagenet}
\begin{center}
\scalebox{0.93}{
\begin{tabular}{l|cc|ccccccc|cc} 
\hline
& \multirow{2}{*}{Teacher} & \multirow{2}{*}{Student} & \multirow{2}{*}{KD} & \multirow{2}{*}{AT$^*$} & \multirow{2}{*}{RKD} & \multirow{2}{*}{SP} & \multirow{2}{*}{CC} & \multirow{2}{*}{AFD$^*$} & \multirow{2}{*}{CRD(+KD)} & \multicolumn{2}{c}{AMD} \\
&&&&&&&&&&(g)&(g+l)\\\hline
Top-1 & 73.31 & 69.75 & 70.66 & 70.70 & 70.59 & 70.79 & 69.96 & 71.38 & 71.17(71.38) & \textbf{71.58} & 71.47 \\
Top-5 & 91.42 & 89.07 & 89.88 & 90.00 & 89.68 & 89.80 & 89.17 & -- & 90.13(90.49) & \textbf{90.50} & 90.49\\
\hline
\end{tabular}
}
\end{center}
\end{table*}


In this section, we explore the performance of attention based distillation approaches with different types of combinations for teacher and student.
We set four types of combinations for teacher and student that consist of the same or different structure of networks. The four types of combinations are described in Table \ref{table:info_settings}. Since the proposed method is relevant to using attention maps, we implemented various baselines that are state-of-the-art attention based distillation methods, including AT \cite{zagoruyko2016paying}, AFDS \cite{wang2019pay}, and AFD \cite{ji2021show}. As described in section \ref{sec:related_work}, AT \cite{zagoruyko2016paying} uses activation-based spatial attention maps for transferring from teacher to student. AFDS \cite{wang2019pay} includes attentive feature distillation and accelerates the transfer-learned model by feature selection. Additional layers are used to calculate a transfer importance predictor used to measure the importance of the source activation maps and enforce a different penalty for training a student. AFD \cite{ji2021show} extracts channel and spatial attention maps and identifies similar features between teacher and student, which are used to control the distillation intensities for all possible pairs and compensate for the limitation of learning to transfer (L2T) \cite{pmlr-v97-jang19b} using manually selected links.
We implemented AFDS \cite{wang2019pay} when the dimension size of features for intermediate layers from the student is the same as the one from the teacher to concentrate on the distillation effects. We use four datasets that have varying degrees of difficulty in a classification problem. These baselines are used in the following experiments as well.

Table \ref{table:att_CIFAR10} presents the accuracy of various knowledge distillation methods for all setups in Table \ref{table:info_settings} on CIFAR-10 dataset. The proposed method, AMD (global+local), has the best performing results in all cases. Table \ref{table:att_CINIC10} describes the CINIC-10 results. In most cases, AMD (global+local) achieves the best results. For experiments on Tiny-ImageNet, as illustrated in Table \ref{table:att_Tiny}, AMD outperforms previous methods, and AMD (global) shows better results in (a) and (b$^b$) setups. For (c$^b$) and (d) setups, AMD (global+local) provides better results. For experiments on ImageNet, standard KD is not applied to baselines and Full KD is utilized. Teacher and student networks are ResNet34 and ResNet18, respectively. The results of baselines are referred from prior works \cite{ji2021show, tian2019contrastive}. As described in Table \ref{test_accuracy_table_on_imagenet}, AMD (global) outperforms other distillation methods, increasing the top-1 and top-5 accuracy by 1.83\% and 1.43\% over the results of learning from scratch, respectively.

\begin{figure}[htb!] 
\includegraphics[scale=0.47]  {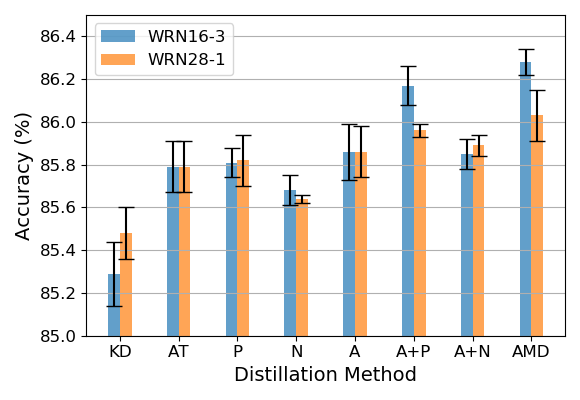} 
\centering
\caption{Accuracy ($\%$) of students (WRN16-1) trained with teachers (WRN16-3 and WRN28-1) on CIFAR-10 for various loss functions.}

\label{figure:amdloss_cifar10}
\end{figure}

Compared to KD, AT obtains better performance in most cases across datasets. That is, the attention map helps the teacher to transfer its knowledge. Even though there is a case that AT shows lower performance than KD in Table \ref{table:att_Tiny}, AMD outperforms KD in all cases. It verifies that applying the discriminative angular distance metric for knowledge distillation maximizes the attention map's efficacy of transferring the knowledge and performs to complement the traditional KD for various combinations of teacher and student. The accuracies of SP with setup (a) and (d), and AFD with setup (d), are even lower than the accuracy of learning from scratch, while AMD performs better than other methods as shown in Table \ref{table:att_Tiny}. When the classification problem is harder, AMD (global) can perform better than AMD (global+local) in some cases. When the teacher and student have different channels or architectural styles, AMD (global+local) can generate a better student than AMD (global).


\textbf{Components of AMD loss function.} As described in Equation \ref{eq:amd}, angular margin distillation loss function ($\mathcal{L}_{AM}(Q_{Tp},Q_{Tn},Q_{Sp},Q_{Sn})$) includes three components (A, P, N). To verify the performance of each component in AMD loss, we experiment with each component separately. As shown in Figure \ref{figure:amdloss_cifar10}, among all components, (A) provides the strongest contribution. Each component in AMD contributes to improvements in performance, which transfers different knowledge. Adding one component to the other one provides richer information, which leads to better performance. The combination of all the components (AMD) show a much higher performance. This result indicates that all components (AMD) are critical to distilling the best student model.

\begin{figure*}[] 
\includegraphics[scale=0.35] {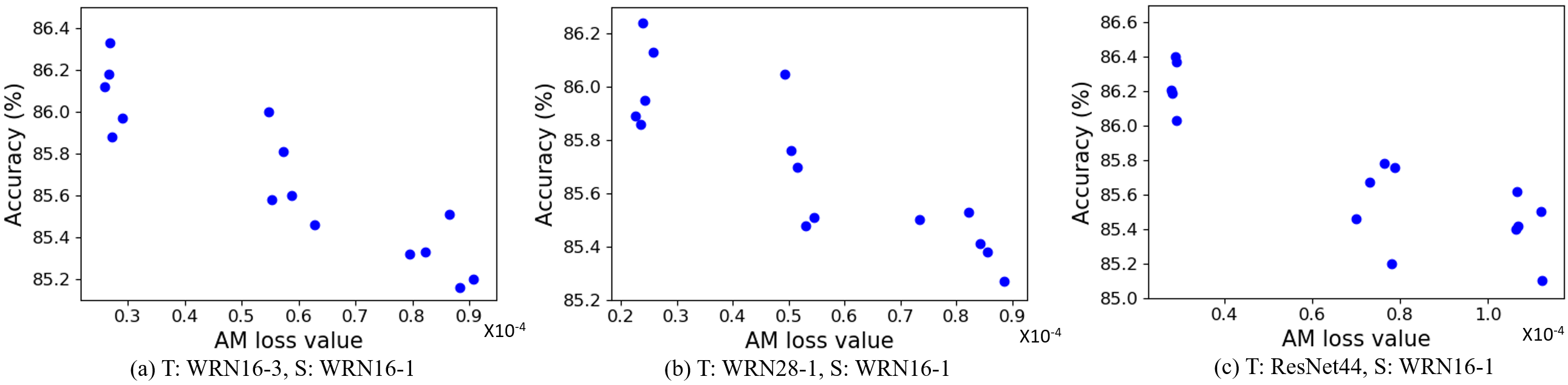}
\centering
\caption{$\mathcal{L_{A}}$ vs. Accuracy ($\%$) for (from left to right) WRN16-1 students (S) trained with WRN16-3, WRN28-1, and ResNet44 teachers (T), on CIFAR-10.}
\label{figure:sample_figure1}
\end{figure*}

\begin{figure*}[] 
\includegraphics[scale=0.5] {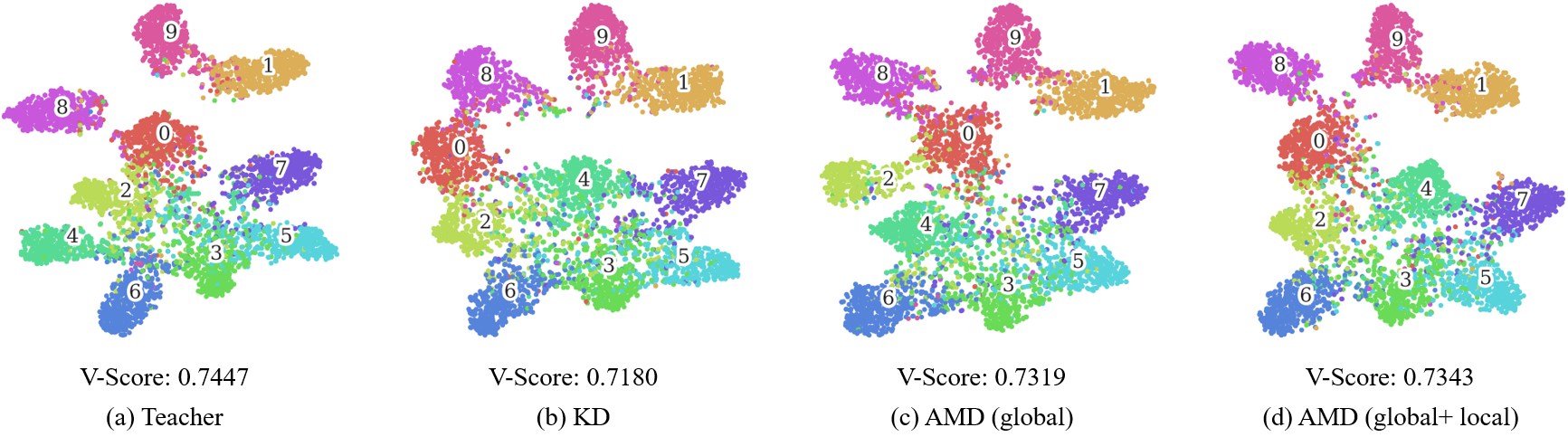}
\centering
\caption{t-SNE plots of output for teacher model (ResNet44) and students (WRN16-1) trained with KD and AMD on CIFAR-10.}
\label{figure:tsne}
\end{figure*}

In Figure \ref{figure:sample_figure1} we show $\mathcal{L_{A}}$ vs. accuracy, when using KD, SP, and AMD (global), for WRN16-1 students trained with WRN16-3, WRN28-1, and ResNet44 teachers, on CIFAR-10 testing set. As shown in Figure \ref{figure:sample_figure1}, when the loss value is smaller, the accuracy is higher. Thus, these plots verify that $\mathcal{L_{A}}$ and performance are correlated.

\textbf{t-SNE visualization and cluster metrics.} To measure the clustering performance, we plot t-SNE \cite{vandermaaten08a} and calculate V-Score \cite{rosenberg2007v} of outputs from penultimate layers of KD and the proposed method on CIFAR-10, where V-Score is clustering metrics implying a higher value is better clustering. As shown in Figure \ref{figure:tsne}, compared to KD, AMD helps get tighter clusters and better separation between classes as seen in higher V-Score.



\subsection{Effect of teacher capacity} \label{sec:various_teacher}

\begin{table*}[]
\centering
\renewcommand{\tabcolsep}{1.2mm} 
\caption{Accuracy ($\%$) with various knowledge distillation methods for different combinations of teachers and students. ``Teacher'' and ``Student'' denote results of the model used to train the distillation methods and trained from scratch, respectively. ``$\mathrm{g}$'' and ``$\mathrm{l}$'' denote using global and local feature distillation, respectively.}

\begin{center}
\scalebox{0.75}{
\begin{tabular}{c |c c| c c| c c c c c c| c c c| c c | c}

\hline
\centering
Method & \multicolumn{4}{c|}{CIFAR-10} &
\multicolumn{9}{c|}{CINIC-10} & \multicolumn{3}{c}{Tiny-ImageNet} \\
\hline

\multirow{4}{*}{Teacher} & WRN & WRN & WRN & WRN &
WRN & WRN & WRN & WRN & WRN & WRN & WRN & WRN & M.Net &
WRN & WRN & WRN\\
& 28-1 & 40-1 & 16-3 & 16-8 &
16-3 & 16-8 & 28-1 & 40-1 & 28-3 & 40-2 & 16-3 & 28-3 & V2 &
 40-1 & 40-2 & 16-3 \\
& (0.4M, & (0.6M, & (1.5M, & (11.0M, &
(1.5M, & (11.0M, & (0.4M, & (0.6M, & (3.3M, & (2.2M, & (1.5M, & (3.3M, & (0.6M, &
(0.6M, & (2.3M, & (1.6M, \\
& 85.84) & 86.39) & 88.15) & 89.50) &
75.65) & 77.97) & 73.91) & 74.49) & 77.14) & 76.66) & 75.65) & 77.14) & 80.98) &
55.28) & 60.18) & 58.78)  \\

\hline
\multirow{3}{*}{Student} & \multicolumn{2}{c|}{WRN16-1} & \multicolumn{2}{c|}{WRN28-1} &
\multicolumn{6}{c|}{WRN16-1} & \multicolumn{3}{c|}{ResNet20} &
\multicolumn{2}{c|}{WRN16-1} & ResNet20 \\
 & \multicolumn{2}{c|}{(0.2M,} & \multicolumn{2}{c|}{(0.4M, } &
\multicolumn{6}{c|}{(0.2M, } & \multicolumn{3}{c|}{(0.3M, } &
\multicolumn{2}{c|}{(0.2M,} & (0.3M, \\
 & \multicolumn{2}{c|}{84.11{\scriptsize$\pm$0.21})} & \multicolumn{2}{c|}{85.59{\scriptsize$\pm$0.13})} &
\multicolumn{6}{c|}{72.05{\scriptsize$\pm$0.12})} & \multicolumn{3}{c|}{72.74{\scriptsize$\pm$0.09})} &
\multicolumn{2}{c|}{49.45{\scriptsize$\pm$0.20})} & 51.75{\scriptsize$\pm$0.19} \\
\hline 
\multirow{2}{*}{KD} & 85.48 & 85.42 & 86.57 & 86.68 &
74.31 & 74.17 & 74.66 & 74.45 & 74.26 & 74.29 & 75.12 & 74.97 & 76.69 &
49.56 & 49.67 & 51.72 \\
 & {\scriptsize$\pm$0.12} & {\scriptsize$\pm$0.11} & {\scriptsize$\pm$0.16} & {\scriptsize$\pm$0.08} &
{\scriptsize$\pm$0.10} & {\scriptsize$\pm$0.16} & {\scriptsize$\pm$0.08} & {\scriptsize$\pm$0.03} & {\scriptsize$\pm$0.06} & {\scriptsize$\pm$0.09} & {\scriptsize$\pm$0.11} & {\scriptsize$\pm$0.07} & {\scriptsize$\pm$0.06} &
{\scriptsize$\pm$0.17} & {\scriptsize$\pm$0.13} & {\scriptsize$\pm$0.13} \\
\multirow{2}{*}{AT} & 85.79 & 85.79 & 86.77 & 87.00 &
74.63 & 74.23 & 74.73 & 74.55 & 74.19 & 74.48 & 75.33 & 75.18 & 77.34 &
49.79 & 49.62 & 51.65 \\
 & {\scriptsize$\pm$0.12} & {\scriptsize$\pm$0.11} & {\scriptsize$\pm$0.11} & {\scriptsize$\pm$0.05} &
{\scriptsize$\pm$0.13} & {\scriptsize$\pm$0.14} & {\scriptsize$\pm$0.02} & {\scriptsize$\pm$0.06} & {\scriptsize$\pm$0.11} & {\scriptsize$\pm$0.08} & {\scriptsize$\pm$0.11} & {\scriptsize$\pm$0.09} & {\scriptsize$\pm$0.10} &
{\scriptsize$\pm$0.22} & {\scriptsize$\pm$0.16} & {\scriptsize$\pm$0.05} \\
\multirow{2}{*}{SP} & 85.77 & 85.90 & 86.56 & 86.94 &
74.43 & 74.34 & 74.94 & 74.86 & 75.04 & 74.81 & 75.29 & 75.50 & 73.71 &
49.89 & 49.59 & 51.87 \\
 & {\scriptsize$\pm$0.07} & {\scriptsize$\pm$0.11} & {\scriptsize$\pm$0.09} & {\scriptsize$\pm$0.08} &
{\scriptsize$\pm$0.11} & {\scriptsize$\pm$0.13} & {\scriptsize$\pm$0.11} & {\scriptsize$\pm$0.07} & {\scriptsize$\pm$0.10} & {\scriptsize$\pm$0.09} & {\scriptsize$\pm$0.10} & {\scriptsize$\pm$0.09} & {\scriptsize$\pm$0.10} &
{\scriptsize$\pm$0.20} & {\scriptsize$\pm$0.25} & {\scriptsize$\pm$0.09} \\

\hline
AMD & 86.04 & 86.03 & 87.13 & 87.22 &
75.04 & 74.93 & 75.14 & \textbf{75.12} & 74.72 & 74.95 & 75.66 & 75.61 & 78.45 &
\textbf{50.15} & 49.88 & 51.89\\
(g) & {\scriptsize$\pm$0.12} & {\scriptsize$\pm$0.09} & {\scriptsize$\pm$0.14} & {\scriptsize$\pm$0.17} &
{\scriptsize$\pm$0.11} & {\scriptsize$\pm$0.09} & {\scriptsize$\pm$0.06} & {\scriptsize$\pm$0.07} & {\scriptsize$\pm$0.07} & {\scriptsize$\pm$0.20} & {\scriptsize$\pm$0.08} & {\scriptsize$\pm$0.06} & {\scriptsize$\pm$0.03} &
{\scriptsize$\pm$0.11} & {\scriptsize$\pm$0.20} & {\scriptsize$\pm$0.25}  \\

AMD & \textbf{86.10} & \textbf{86.15} & \textbf{87.35} & \textbf{87.31} &
\textbf{75.18} & \textbf{75.20} & \textbf{75.21} & 75.10 & \textbf{75.22} & \textbf{75.04} & \textbf{75.75} & \textbf{75.76} & \textbf{78.62} &
49.97 & \textbf{50.07} & \textbf{52.12} \\
(g+l) & {\scriptsize$\pm$0.10} & {\scriptsize$\pm$0.06} & {\scriptsize$\pm$0.10} & {\scriptsize$\pm$0.15} &
{\scriptsize$\pm$0.09} & {\scriptsize$\pm$0.05} & {\scriptsize$\pm$0.04} & {\scriptsize$\pm$0.04} & {\scriptsize$\pm$0.07} & {\scriptsize$\pm$0.06} & {\scriptsize$\pm$0.08} & {\scriptsize$\pm$0.11} & {\scriptsize$\pm$0.04} &
{\scriptsize$\pm$0.18} & {\scriptsize$\pm$0.10} & {\scriptsize$\pm$0.15} \\
\hline

\end{tabular}
}
\end{center}
\label{table:various_CapacityT}
\end{table*}

To understand the effect of the capacity of the teacher, we implemented various combinations of teacher and student, where the teacher has a different capacity. We use well-known benchmarks for image classification which are WRN \cite{zagoruyko2016wide}, ResNet \cite{he2016deep}, and MobileNetV2 (M.NetV2) \cite{sandler2018mobilenetv2}. We applied the same settings as in the experiments of the previous section.

The results in classification accuracy for the student models are described in Table \ref{table:various_CapacityT} across three datasets, trained with attention based and non-attention based methods  \cite{hinton2015distilling, zagoruyko2016paying, tung2019similarity}. The number of trainable parameters are noted in in brackets. For all cases, the proposed method, AMD, shows the highest accuracy. When the complexity of the dataset is higher and the depth of teacher is largely different from the one of the student, AMD (global) tends to generate a better student than AMD (global+local). When a larger capacity of students is used, the accuracy observed is higher. This is seen in the results from WRN16-1 and ResNet20 students with WRN16-3 and WRN28-3 teachers on CINIC-10 dataset. For the combinations, ResNet20 students having a larger capacity than WRN16-1 generate better results. Furthermore, on CIFAR-10, when a WRN16-3 teacher is used, a WRN28-1 student achieves 87.35$\%$ for AMD (global+local), whereas a WRN16-1 student achieves 86.36$\%$ for AMD (global+local). On Tiny-ImageNet, when AMD (global+local) is used, the accuracy of a ResNet20 student is 52.12$\%$, which is higher than the accuracy of a WRN16-1 student, which is 49.92$\%$. 

Compared to KD, in most cases, AT achieves better performance. However, when the classification problem is difficult, such as when using Tiny-ImageNet, and when WRN40-2 teacher and WRN16-1 student are used, both AT and SP show worse performance than KD. When the WRN16-3 teacher and ResNet20 student are used, KD and AT perform worse than the model trained from scratch. The result of AT is even lower than that of KD. So, there are cases where AT and SP cannot complement the performance of the traditional KD. On the other hand, for the proposed method, the results are better than the baselines in all the cases. Interestingly, on CIFAR-10 and CINIC-10, the result of a WRN16-1 student trained by AMD with a WRN28-1 teacher is even better than the result of the teacher. Therefore, we conclude that the proposed method maximizes the attention map's efficacy of transferring the knowledge and complements traditional KD.

Also, when applying the larger teacher model and the smaller student model, the performance degradation of AMD can occur. 
For example, on CINIC-10, WRN16-1 student trained with WRN40-1 (0.6M) teacher outperforms the one trained with WRN40-2 (2.3M) teacher.
Both AMD and other methods produce some cases with lower performance when a better (usually larger) teacher is used. This is consistent with prior findings \cite{cho2019efficacy, wang2021knowledge, stanton2021does} that a better teacher does not always guarantee a better student.

\textbf{Heterogeneous teacher-student.}
In Table \ref{table:various_CapacityT}, we present the results of the teacher-student combinations from similar architecture styles. Tian \emph{et al.} \cite{tian2019contrastive} found that feature distillation methods such as SP sometimes struggled to find the optimal solution in different architecture styles. 
In this regard, we implemented heterogeneous teacher-student combination, where the teacher and student have very different structure of networks. We use vgg \cite{simonyan2014very} network to compose heterogeneous combinations.  

As describe in Table \ref{table:heteroT}, we observe similar findings, showing degraded performance in using SP when vgg13 teacher and ResNet20 student are used, while AMD consistently outperforms all baselines we explored. Also, in most cases, WRN16-8 teacher distills a better student (vgg8) than WRN28-1 teacher. However, KD and SP shows better performance with WRN28-1 teacher, which corroborates a better teacher does not always distill a better student.

\begin{table}[htb!]
\centering
\renewcommand{\tabcolsep}{1.2mm} 
\caption{Accuracy ($\%$) with various knowledge distillation methods for different structure of teachers and students on CIFAR-10. ``Teacher'' and ``Student'' denote results of the model used to train the distillation methods and trained from scratch, respectively. ``$\mathrm{g}$'' and ``$\mathrm{l}$'' denote using global and local feature distillation, respectively.}

\begin{center}
\scalebox{0.9}{
\begin{tabular}{c |c c| c c }

\hline
\centering

\multirow{4}{*}{Teacher} & WRN & WRN & \multirow{2}{*}{vgg13} & M.Net \\
& 28-1 & 16-8 & & V2 \\
& (0.4M, & (11.0M, & (9.4M, & (0.6M,  \\
& 85.84) & 89.50) & 88.56) & 89.61)  \\

\hline
\multirow{3}{*}{Student} & \multicolumn{2}{c|}{vgg8} & ResNet20 & ResNet26  \\
 & \multicolumn{2}{c|}{(3.9M,} & (0.3M, & (0.4M, \\
 & \multicolumn{2}{c|}{85.41{\scriptsize$\pm$0.06})} & 85.20{\scriptsize$\pm$0.17}) & 85.65{\scriptsize$\pm$0.20}) \\
\hline 
\multirow{2}{*}{KD} & 86.93 & 86.74 & 85.39 & 87.74\\
 & {\scriptsize$\pm$0.11} & {\scriptsize$\pm$0.13} & {\scriptsize$\pm$0.07} & {\scriptsize$\pm$0.08} \\
\multirow{2}{*}{AT} & 87.16 & 87.29 & 85.63 & 88.61 \\
 & {\scriptsize$\pm$0.09} & {\scriptsize$\pm$0.10} & {\scriptsize$\pm$0.20} & {\scriptsize$\pm$0.04}\\
\multirow{2}{*}{SP} & 87.29 & 86.82 & 85.00 & 85.78 \\
 & {\scriptsize$\pm$0.00} & {\scriptsize$\pm$0.07} & {\scriptsize$\pm$0.07} & {\scriptsize$\pm$0.10} \\

\hline
AMD & 87.43 & 87.61 & 86.18 & \textbf{88.70}\\
(g) & {\scriptsize$\pm$0.04} & {\scriptsize$\pm$0.11} & {\scriptsize$\pm$0.14} & {\scriptsize$\pm$0.03} \\

AMD & \textbf{87.56} & \textbf{87.63} & \textbf{86.41} & 88.42 \\
(g+l) & {\scriptsize$\pm$0.03} & {\scriptsize$\pm$0.07} & {\scriptsize$\pm$0.04} & {\scriptsize$\pm$0.08} \\
\hline

\end{tabular}
}
\end{center}
\label{table:heteroT}
\end{table}


\subsection{Ablations and sensitivity analysis} \label{sec:Param}

In this section, we investigate sensitivity for hyperparameters ($\gamma$ and $m$) used for the angular margin based attention distillation. 

\subsubsection{\makebox{Effect of angular distillation hyperparameter $\gamma$}}
\begin{figure}[htb!] 
\includegraphics[scale=0.315] {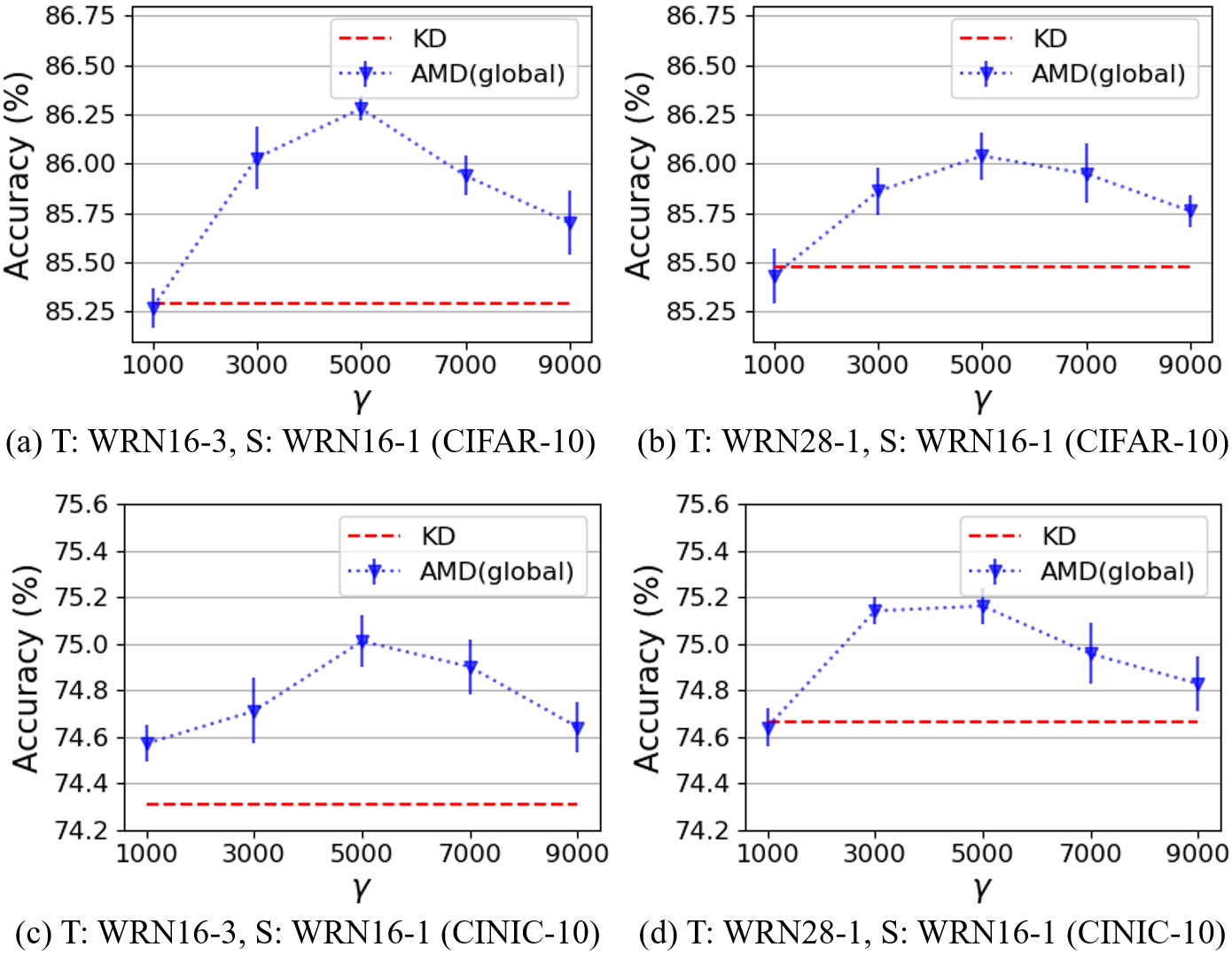}
\centering
\caption{Accuracy ($\%$) of students (WRN16-1) for AMD (global) with various $\gamma$, trained with teachers (WRN16-3 and WRN28-1) on CIFAR-10 and CINIC-10. ``T'' and ``S'' denote teacher and student, respectively.}
\label{figure:cifar10_gamma}
\end{figure}

The results of a student model (WRN16-1) for AMD (global) trained with teachers (WRN16-3 and WRN28-1) by using various $\gamma$ on CIFAR-10 (the first row) and CINIC-10 (the second row) are depicted in Figure \ref{figure:cifar10_gamma} ($m$ = 1.35). When $\gamma$ is 5000, all results show the best accuracy. For CIFAR-10, when WRN16-3 is used as a teacher, the accuracy of $\gamma$ = 3000 is higher than that of $\gamma$ = 7000. However, for WRN28-1 as a teacher, the accuracy of $\gamma$ = 7000 is higher than that of $\gamma$ = 3000. When $\gamma$ is 1000, the accuracy is lower than KD, implying that it does not complement KD and adversely affects the performance. On the other hand, for CINIC-10, when the WRN16-3 teacher is used, the result of $\gamma$ = 7000 is better than that of $\gamma$ = 3000. But, for the WRN28-1 teacher, $\gamma$ = 3000 is higher than that of $\gamma$ = 7000. Therefore, $\gamma$ values between 3000 and 7000 achieve good performance, while too small or large $\gamma$ values do not help much with improvement. Therefore, setting the proper $\gamma$ value is important for performance. We recommend using $\gamma$ as 5000, which produces the best results across datasets and combinations of teacher and student.

\begin{figure}[htb!] 
\includegraphics[scale=0.29] {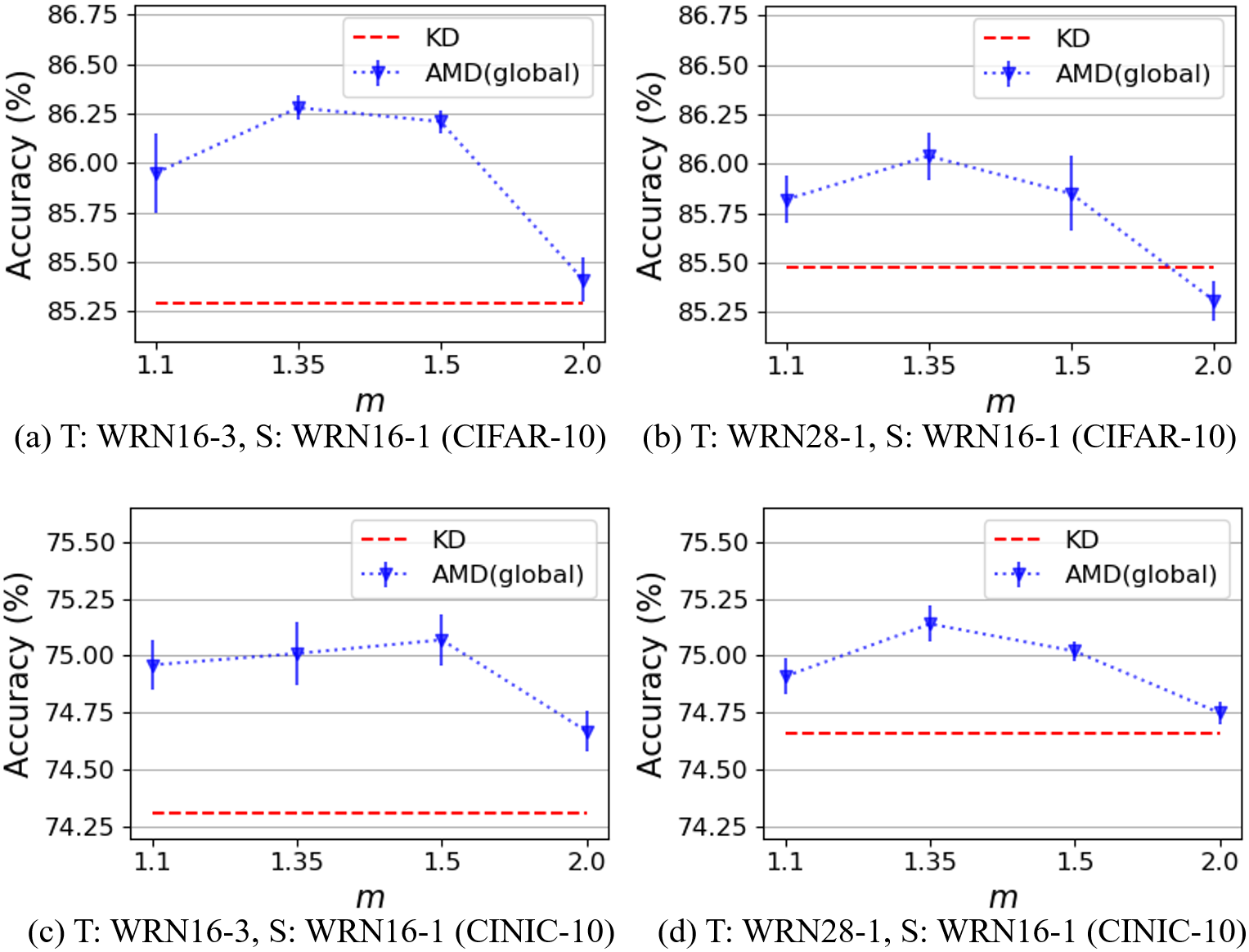} 
\centering
\caption{Accuracy ($\%$) of students (WRN16-1) for AMD (global) with various angular margin $m$, trained with teachers (WRN16-3 and WRN28-1) on CIFAR-10 and CINIC-10. ``T'' and ``S'' denote teacher and student, respectively.}
\label{figure:sample_margin}
\end{figure}

\subsubsection{Effect of angular margin $m$}

The results of a student model (WRN16-1) for AMD (global) trained with teachers (WRN16-3 and WRN28-1) by various angular margin $m$ on CIFAR-10 (the first row) and CINIC-10 (the second row) are illustrated in Figure \ref{figure:sample_margin} ($\gamma$ = 5000).
As described in section \ref{proposed_AMD}, using the large value of $m$ corresponds to producing more distinct positive features in the attention map and making a large gap between positive and negative features for distillation. When $m$ is 1.35 for the WRN16-3 teacher, the WRN16-1 student shows the best performance of 86.28$\%$ on CIFAR-10. When $m$ = 1.5 for CINIC-10, the student's accuracy is 75.13$\%$, which is higher than when $m$ = 1.35. When the teacher is WRN28-1, the student produces the best accuracy with $m$ = 1.35 on both datasets. The student model with $m$ = 1.35 performs better than the one with $m$ = 1.1 and 2.0. When the complexity of the dataset is higher, using $m$ (1.5) which is larger than 1.35 can produce a good performance.
When $m$ = 1.0 (no additional margin applied to the positive feature) for CIFAR-10 and CINIC-10 with setup ($b$), the results are 85.81$\%$ and 74.83$\%$, which are better than those of 85.31$\%$ and 74.75$\%$ from $m$ = 2.0, respectively.
This result indicates that it is important to set an appropriate $m$ value for our method. We believe that angular margin plays a key role in determining the gap between positive and negative features. As angular margin increases, the positive features are further emphasized, and in this case of over-emphasis by a much larger $m$, the performance is worse than that of the smaller $m$.
We recommend using a margin $m$ of around 1.35 ($m > 1.0$), which generates the best results in most cases.

\subsection{Analysis with activation maps} \label{sec:activation_map}
To analyze results with intermediate layers, we adopt Grad-CAM \cite{selvaraju2017grad} which uses class-specific gradient information to visualize the coarse localization map of the important regions in the image. In this section, we present the activation maps from intermediate layers and the high level of the layer with various methods. The red region is more crucial for the model prediction than the blue one.

\begin{figure}[htb!] 
\includegraphics[width=0.4\textwidth] {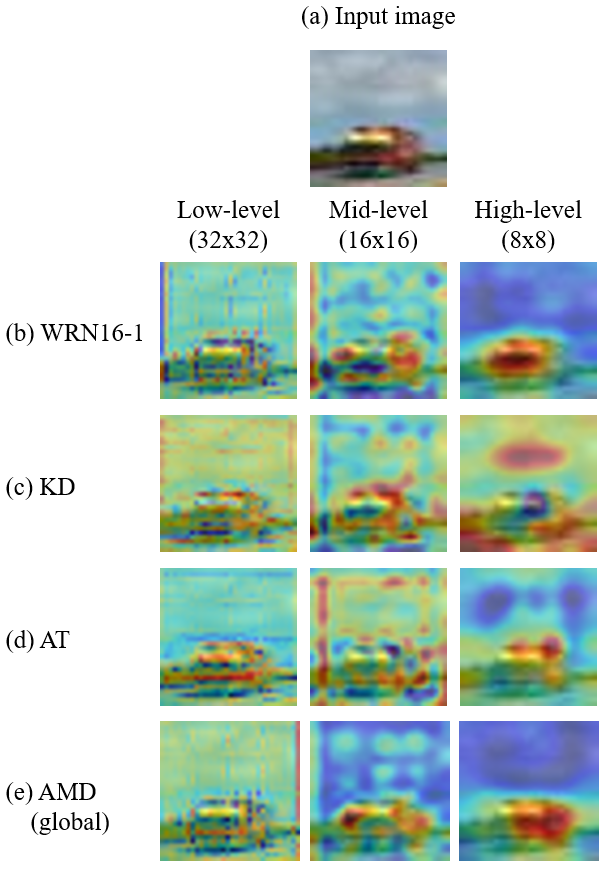} 
\centering
\caption{Activation maps for different levels of students (WRN16-1) trained with a teacher (WRN16-3) on CIFAR-10.}
\label{figure:activation_levels}
\end{figure}

\subsubsection{Activation maps for the different levels of layers}

\begin{figure}[htb!] 
\includegraphics[width=0.43\textwidth] {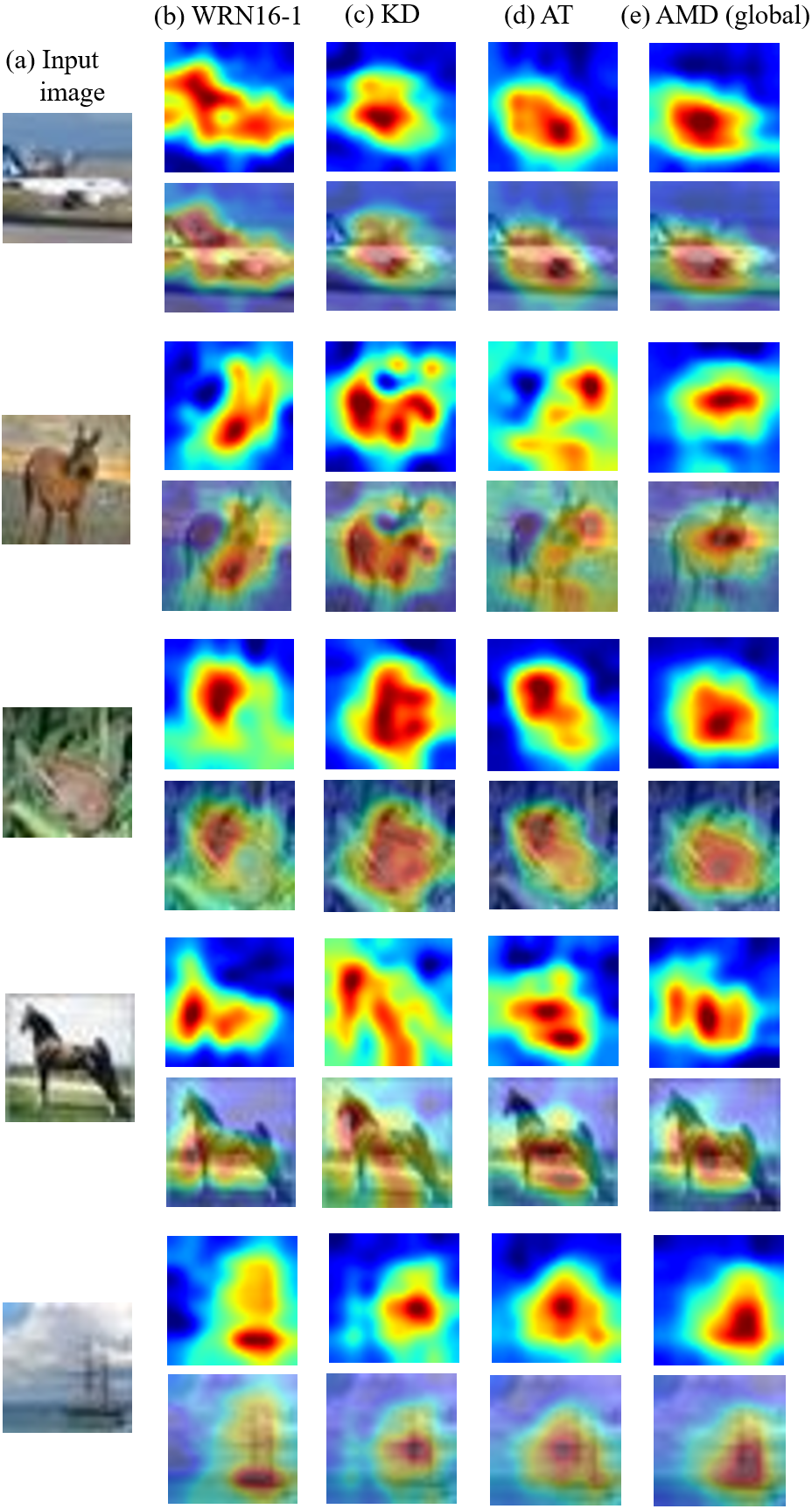}
\centering
\caption{Activation maps of high-level from students (WRN16-1) trained with a teacher (WRN16-3) for different input images on CIFAR-10.}
\label{figure:activation}
\end{figure}

The activation maps from intermediate layers with various methods are shown in Figure \ref{figure:activation_levels}. The proposed method, AMD, shows intuitively similar activated regions to the traditional KD \cite{hinton2015distilling} in the low-level. However, at mid-level and high-level, the proposed method represents the higher activations around the region of a target object, which is different from the previous methods \cite{hinton2015distilling, zagoruyko2016paying}. Thus, the proposed method can classify positive and negative areas more discriminatively, compared to the previous methods \cite{hinton2015distilling, zagoruyko2016paying}.
The high-level activation maps with various input images are described in Figure \ref{figure:activation}. The activation from proposed method is seen to be more centered on the target. The result shows that the proposed method performs better in focusing on the foreground object distinctly with high weight, while being less distracted by the background compared to other methods \cite{hinton2015distilling, zagoruyko2016paying}. With higher weight over regions of interest, the student from the proposed method has a stronger discrimination ability. Therefore, the proposed method guides student models to increase class separability.


\subsubsection{Activation maps for global and local distillation of AMD}

\begin{figure}[htb!] 
\includegraphics[width=0.43\textwidth] {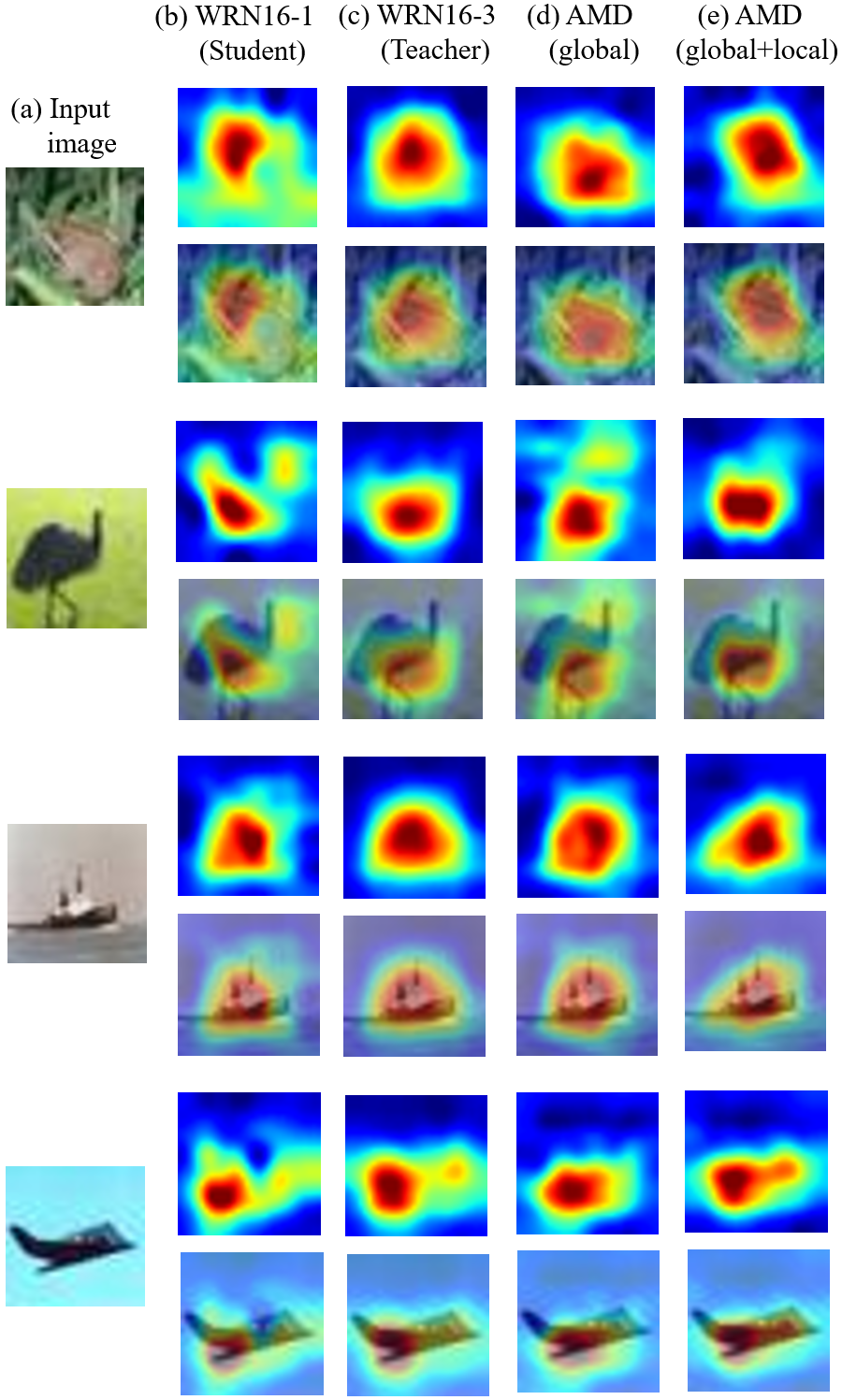}
\centering
\caption{Activation maps of high-level from students (WRN16-1) for AMD trained with a teacher (WRN16-3) for different input images on CIFAR-10.}
\label{figure:attention_AMD}
\end{figure}

To investigate the impact of using global and local features for AMD, we illustrate relevant results in Figure \ref{figure:attention_AMD}. When both global and local features are used for distillation, the activated area is located and shaped more similar to the teacher, than using the global feature only. Also, AMD (global+local) focuses more on the foreground object with higher weights than AMD (global). AMD (global+local) guides the student to focus more on the target regions and finds discriminative regions. Thus, using global and local features is better than using global features alone for the proposed method.

\subsection{Combinations with existing methods} \label{sec:combi_methods}
Even if a model shows good performance in classification, it may have miscalibration problems \cite{guo2017calibration} and may not always obtain improved results from combining with other robust methods. In this section, to evaluate the generalizability of models trained by each method and to explore if the method can complement other methods, we implement experiments with various existing methods. 
We use the method in various ways to demonstrate how easily it can be combined with any previous learning tasks. We trained students with fine-grained features \cite{wang2019distilling, wang2020fully}, augmentation methods, and one of the baselines such as SP \cite{tung2019similarity} that is not based on the attention feature based KD. WRN16-1 students were trained with WRN16-3 and WRN28-1 teachers. We examine whether the proposed method can be combined with other techniques and compare the results to baselines.



\subsubsection{Fine-grained feature-based distillation}

\begin{figure*}[htb!] 
\includegraphics[scale=0.44] {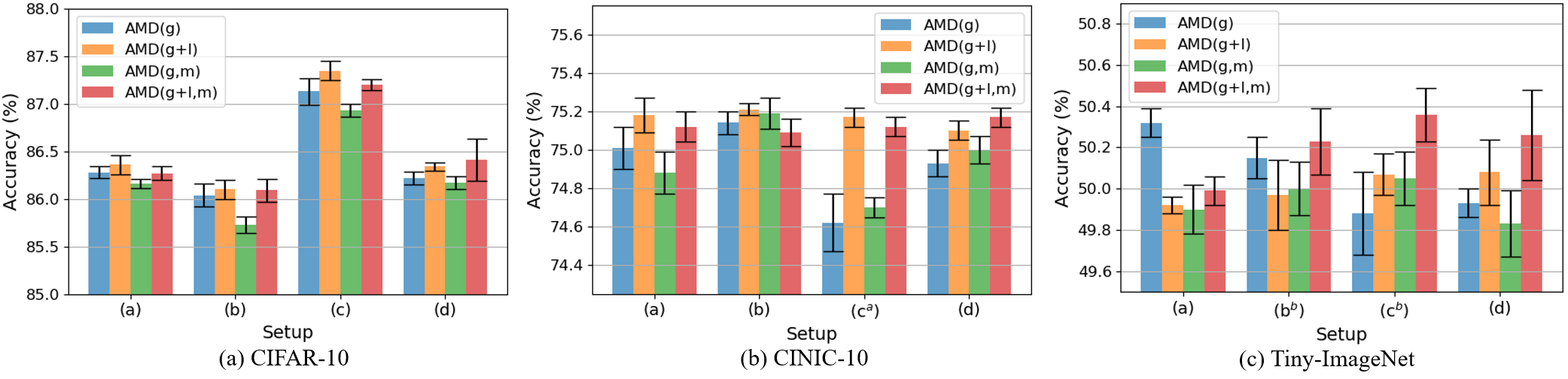}
\centering
\caption{Accuracy ($\%$) from students (WRN16-1) for AMD trained with a teacher (WRN16-3) with/without masked features. ``g'', ``l'', and ``m'' denote global, local, and masked feature, respectively.}
\label{figure:masked}
\end{figure*}

If the features of teacher and student are compatible, it results in a student achieving `minor gains' \cite{wang2019distilling}. To perform better distillation and to overcome the problem of learning minor gains, a technique for generating a fine-grained feature has been used \cite{wang2019distilling, wang2020fully}. For distillation with AMD and creating the fine-grained (masked) feature, a binary mask is adopted when the negative feature is created. For example, if the probability of the point for the negative map is higher than 0.5, the point is multiplied by 1, otherwise by 0. 
Then, compared to non-masking, it boosts the difference between teacher and student, where the difference can be more focused on loss function for training. The results for AMD with or without using masked feature-based distillation are presented in Figure \ref{figure:masked}. The parameter $\gamma$ for training a student based on AMD without masked features is 5000 for all setups across datasets. When masked features are used for AMD, to generate the best results, $\gamma$ of 3000 is applied to setup (b) on CIFAR-10, setup ($c^a$) on CINIC-10, and all setups on Tiny-ImageNet. For CIFAR-10, AMD (global+local) without masked features has the best performing result in most cases. AMD (global+local) with masked features shows the best with setup (d). For CINIC-10, the results of AMD with masked features for setup (d) show the best. For Tiny-ImageNet, in most cases, AMD with masked features performs the best. Therefore, when the complexity of a dataset is high, fine-grained features can help more effectively improve the performance, and the smaller parameter of $\gamma$, 3000, generates better accuracy. Also, AMD (global+local) with masked features produces better performance than AMD (global) with the one. For setup (d) -- different architectures for teacher and student -- with/without masked features, AMD (global+local) outperforms AMD (global). This could be due to the fact that the teacher's features differ from the student's because the two networks have different architectures, resulting in different distributions. So, masked features with both global and local distillation influence more on setup (d) than other setups. The difference between AMD (global) and AMD (global+local) with masked features is also discriminatively shown with the harder problem in classification. If the student's and teacher's architectural styles are similar, the student is more likely to achieve plausible results \cite{wang2021knowledge}.

\subsubsection{Applying augmentation methods}
In this section, we investigate of the compatibility with different types of augmentation methods.

\textbf{Mixup.} Mixup \cite{zhang2017mixup} is one of the most commonly used augmentation methods. We demonstrate here that AMD complements Mixup. Mixup's parameter is set to $\alpha_\text{Mixup}=0.2$. A teacher is trained with the original training set and learns from scratch. A student is trained with Mixup and the teacher model is implemented as a pre-trained model.

\begin{figure}[htb!] 
\includegraphics[width = 0.34\textwidth]
{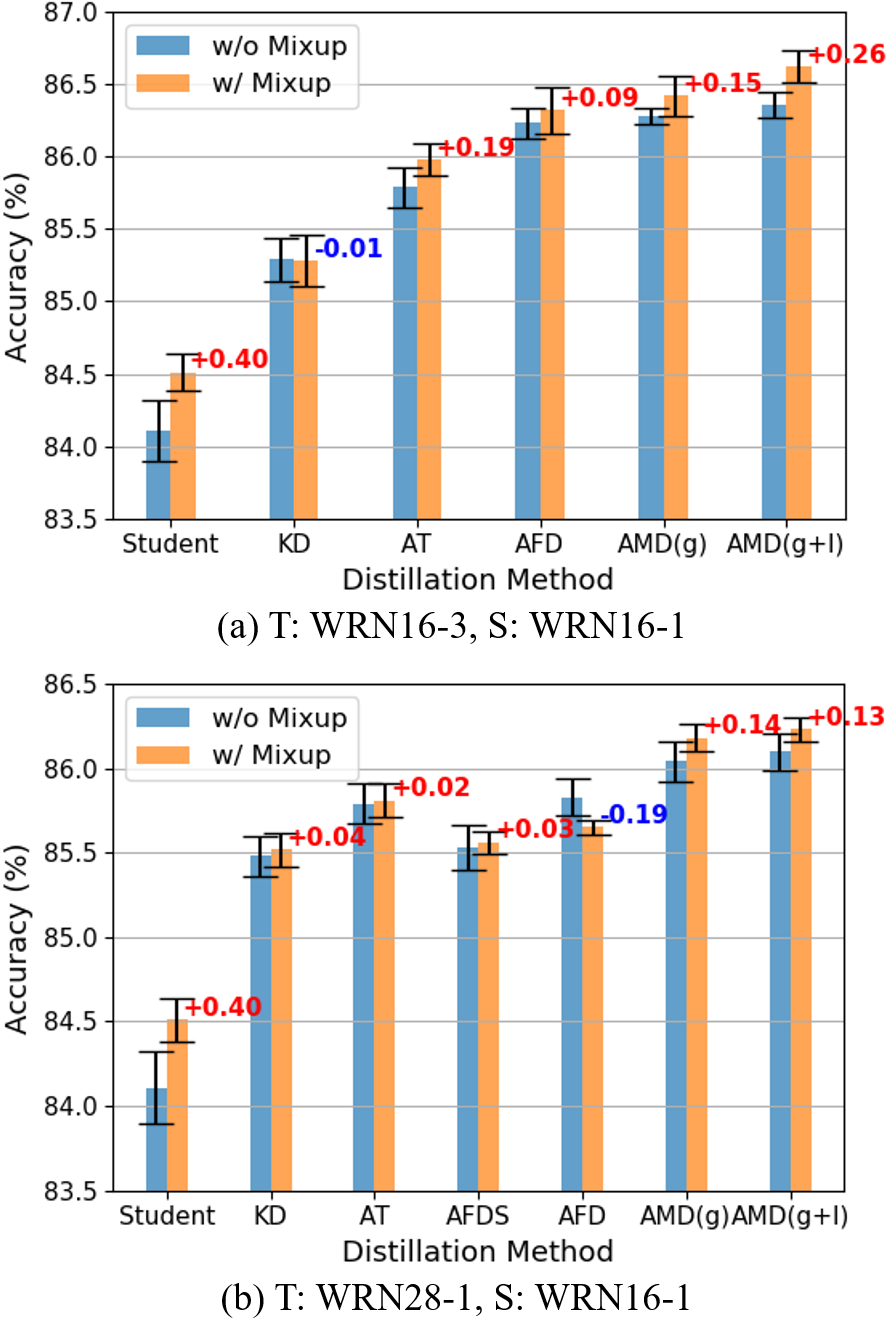} 
\centering
\caption{Accuracy ($\%$) of students (WRN16-1) for knowledge distillation methods, trained with Mixup and a teacher (WRN16-3) on CIFAR-10. ``T'' and ``S'' denote teacher and student, respectively. ``g'' and ``l'' denote using global and local feature distillation, respectively. ``Student'' is a result of WRN16-1 trained from scratch.}
\label{figure:mixup_results}
\end{figure}

\begin{table}[htb!]
\centering
\caption{ECE ($\%$) and NLL ($\%$) for various knowledge distillation methods with Mixup on CIFAR-10. ``$\mathrm{g}$'' and ``$\mathrm{l}$'' denote using global and local feature distillation, respectively. The results (ECE, NLL) for WRN16-3 and WRN28-1 teachers are (1.469$\%$, 44.42$\%$) and (2.108$\%$, 64.38$\%$), respectively. }

\begin{center}
\scalebox{0.72}{
\begin{tabular}{c |c |c c | c c}

\hline
\centering

\multirow{2}{*}{Setup} & \multirow{2}{*}{Method} & \multicolumn{2}{c|}{w/o Mixup }&  \multicolumn{2}{c}{w/ Mixup}\\ 
& & ECE & NLL & ECE & NLL \\
\hline

 & Student & 2.273 & 70.49 & 7.374 (\textcolor{red}{+5.101}) & 90.58 (\textcolor{red}{+20.09}) \\
\hline

\multirow{5}{*}{(a)} & KD \cite{hinton2015distilling} & 2.065 & 63.34 & 1.818 (\textcolor{blue}{-0.247}) & 55.62 (\textcolor{blue}{-7.71}) \\

&AT \cite{zagoruyko2016paying}& 1.978 & 60.48 & 1.652 (\textcolor{blue}{-0.326}) & 50.84 (\textcolor{blue}{-9.64}) \\

&AFD \cite{ji2021show}& \textbf{1.890} & \textbf{56.71} & 1.651 (\textcolor{blue}{-0.240}) & 50.22 (\textcolor{blue}{-6.49})  \\

&AMD (g) & 1.933 & 59.67 & 1.645 (\textcolor{blue}{-0.288}) & 50.33 (\textcolor{blue}{-9.34}) \\

&AMD (g+l) & 1.895 & 57.60 & \textbf{1.592} (\textcolor{blue}{-0.304}) & \textbf{49.68} (\textcolor{blue}{-7.92}) \\

\hline
 
\multirow{6}{*}{(b)} & KD \cite{hinton2015distilling}& 2.201 & 68.75 & 1.953 (\textcolor{blue}{-0.249}) & 58.81 (\textcolor{blue}{-9.93}) \\

&AT  \cite{zagoruyko2016paying}& 2.156 & 67.14 & 1.895 (\textcolor{blue}{-0.261}) & 56.51 (\textcolor{blue}{-10.62}) \\

&AFDS \cite{wang2019pay}& 2.197 & 68.53 & 1.978 (\textcolor{blue}{-0.219}) & 58.86 (\textcolor{blue}{-9.68})  \\

&AFD \cite{ji2021show}& 2.143 & 66.05 & 1.900 (\textcolor{blue}{-0.243}) & 57.68 (\textcolor{blue}{-8.37})  \\

&AMD (g) & \textbf{2.117} & \textbf{66.47} & 1.869 (\textcolor{blue}{-0.248}) & 56.05 (\textcolor{blue}{-10.42}) \\

&AMD (g+l) & 2.123 & 67.51 & \textbf{1.853} (\textcolor{blue}{-0.270}) & \textbf{55.15} (\textcolor{blue}{-12.36}) \\

\hline

\end{tabular} }
\end{center}

\label{table:mixup_result}
\end{table}

\begin{figure}[htb!] 
\includegraphics[scale=0.29] {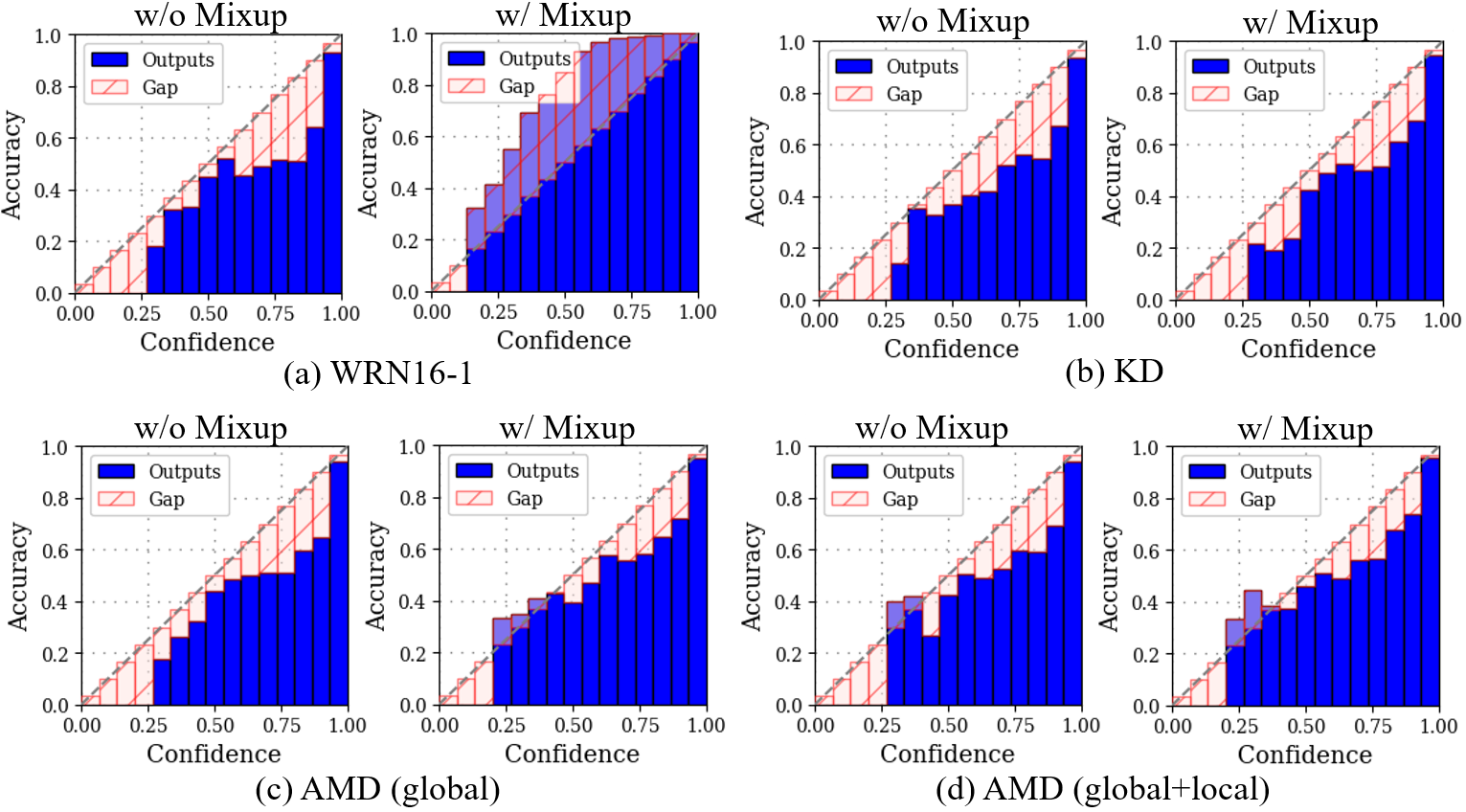} 
\centering
\caption{Reliability diagrams of students (WRN16-1) for knowledge distillation methods, trained with Mixup and a teacher (WRN16-3) on CIFAR-10. For the results of each method, the left is the result without Mixup, and the right is with Mixup.}
\label{figure:mixup_reliability}
\end{figure}


As described in Figure \ref{figure:mixup_results}, with Mixup, most of the methods generate better results. However, KD shows slight degradation when a WRN16-3 teacher is used. This degradation might be related to the artificially blended labels by Mixup. Conventional KD achieves the success by transferring concise logit knowledge. However, with Mixup in KD, the knowledge from a teacher is affected by the mixed labels and is not concise logits, which can hurt distillation quality \cite{das2020empirical}. So, the knowledge for separating different classes can be better encoded by traditional KD (without Mixup) \cite{das2020empirical}. Even though the KD performs degradation with Mixup, all other baselines and proposed methods transferring features with intermediate layers show improvement. Thus, the feature based distillation methods help to reduce the negative effects from noisy logits.
When a WRN28-1 teacher is used, the performance of the student from AFD is degraded. AFD utilizes similarity of features for all possible pairs of the teacher and student. For this combination, Mixup produces noisy features, which can affect to mismatch the pair for distillation to perform degradation. Compared to the baselines, AMD obtains more gains from Mixup. To study the generalizability and regularization effects of Mixup, we measured expected calibration error (ECE) \cite{guo2017calibration, naeini2015obtaining} and negative log likelihood (NLL) \cite{guo2017calibration} for each method. ECE is a metric to measure calibration, representing the reliability of the model \cite{guo2017calibration}. A probabilistic model's quality can be measured by using NLL \cite{guo2017calibration}. The results of training from scratch with Mixup show a higher ECE and NLL than the results of training without Mixup, as seen in Table \ref{table:mixup_result}. However, the methods, including knowledge distillation, generate lower ECE and NLL. This implies that knowledge distillation from teacher to student influences the generation of a better model not only for accuracy but also for reliability. In both (a) and (b), with Mixup, AMD (global+local) shows robust calibration performance. Therefore, we confirm that an augmentation method such as Mixup gets the benefits from AMD in generating better calibrated performance. As can be seen in Figure \ref{figure:mixup_reliability}, WRN16-1 trained from scratch with Mixup produces underconfident predictions \cite{zhang2017mixup}, compared to KD \cite{hinton2015distilling} with Mixup. AMD (global+local) with Mixup achieves the best calibration performance. These results support the advantage of AMD, that it can be easily combined with common augmentation methods to improve the performance in classification with good calibration.

\begin{figure}[htb!] 
\includegraphics[width = 0.4\textwidth]
{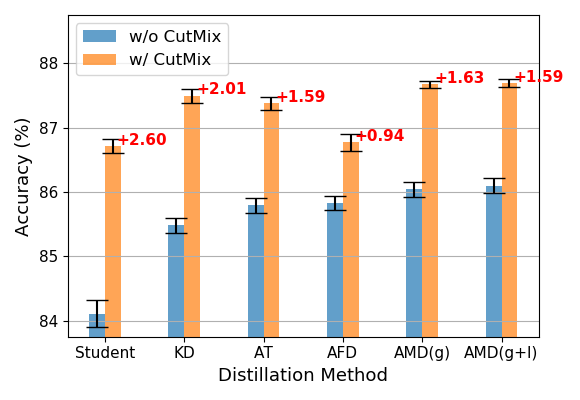} 
\centering
\caption{Accuracy ($\%$) of students (WRN16-1) for knowledge distillation methods, trained with CutMix and a teacher (WRN28-1) on CIFAR-10. ``g'' and ``l'' denote using global and local feature distillation, respectively. ``Student'' is a result of WRN16-1 trained from scratch.}
\label{figure:cutmix_results}
\end{figure}

\textbf{CutMix.} CutMix \cite{yun2019cutmix} one of the most popular augmentation methods, which is more advanced method to Mixup. We evaluate AMD with CutMix. We referred to the previous study to set the parameters for CutMix \cite{yun2019cutmix}.
As illustrated in Figure \ref{figure:cutmix_results}, all methods are improved by CutMix. Compared to other baselines, AFD gains less improvement. Both AMD (global) and AMD (global+local) perform better with CutMix and these results also show that the proposed method can be easily combined with the advanced augmentation methods.


\textbf{MoEx.} To test with a latent space augmentation method, MoEx \cite{li2021feature} is adopted to train with AMD, which is one of the state-of-the-art technique for augmentation. We applied the same parameter by referring to the prior study \cite{li2021feature}. We apply MoEx to a layer before stage 3 in the student network (WRN16-1), which achieves the best with KD.


\begin{figure}[htb!] 
\includegraphics[width = 0.4\textwidth]
{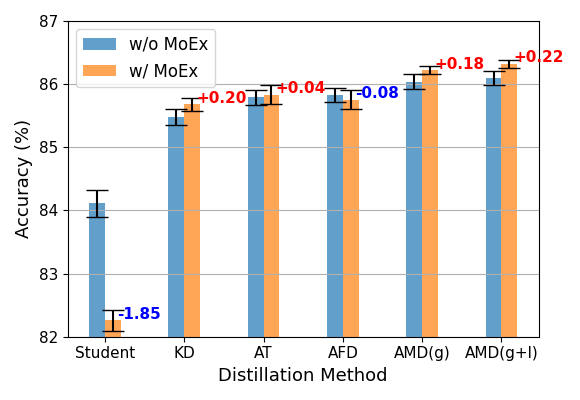} 
\centering
\caption{Accuracy ($\%$) of students (WRN16-1) for knowledge distillation methods, trained with MoEx and a teacher (WRN28-1) on CIFAR-10. ``g'' and ``l'' denote using global and local feature distillation, respectively. ``Student'' is a result of WRN16-1 trained from scratch.}
\label{figure:moex_results}
\end{figure}

As shown in Figure \ref{figure:moex_results}, most of KD based methods with MoEx perform better than the one without MoEx. AFD shows degradation. Since AFD transfers the knowledge considering all pair of features from teacher and student, MoEx in AFD hinders the pair matching and transferring the high quality knowledge. Both AMD (global) and AMD (global+local) outperform baselines. This results verify that latent space augmentation based methods can be combined with the proposed method. Therefore, the proposed method can implement with various augmentation methods to improve the performance.


\begin{figure}[htb!] 
\includegraphics[width = 0.38\textwidth]
{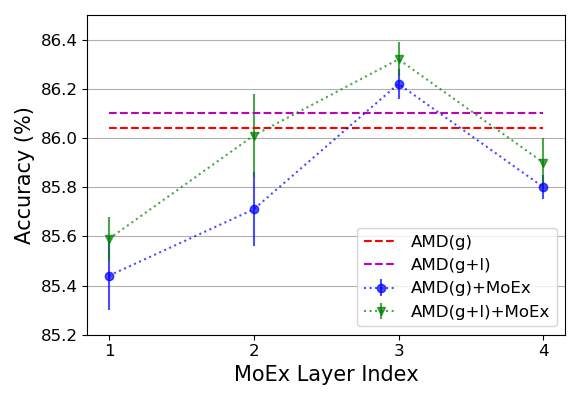} 
\centering
\caption{Accuracy ($\%$) of students (WRN16-1) for knowledge distillation methods, trained with MoEx and a teacher (WRN28-1) on CIFAR-10. We denote the layer index to apply MoEx as (1=before stage 1, 2=before stage 2, 3=before stage 3, 4=after stage 3). ``g'' and ``l'' denote using global and local feature distillation, respectively.}
\label{figure:moex_layer_results}
\end{figure}

Additionally, we explore the work of MoEx at different layers. As described in Figure \ref{figure:moex_layer_results}, when MoEx is applied the layer before stage 3 of the student model, AMD shows the best performance. KD also shows its best when MoEx is applied to a layer before stage 3. This aspect is different from the result of learning from scratch, which shows the best when MoEx is applied to a layer before stage 1 \cite{li2021feature}. Thus, when latent space augmentation is combined with KD based method including baselines and the proposed method, a layer to apply augmentation method has to be chosen considerably. And, these results imply that a layer before stage 3 plays a key role for knowledge distillation.

\subsubsection{Combination with other distillation methods}

\begin{figure}[htb!] 
\includegraphics[width = 0.33\textwidth] {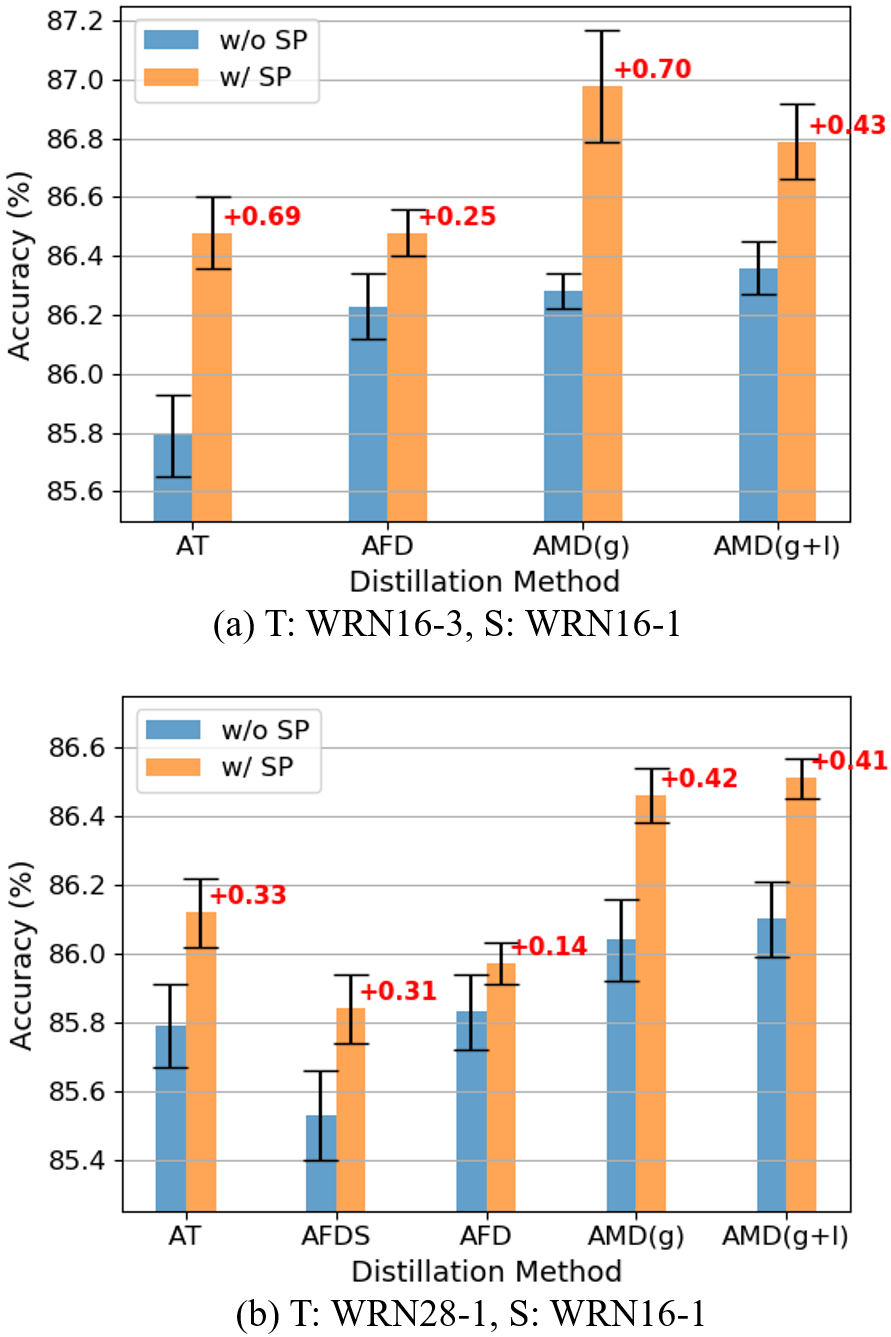} 
\centering
\caption{Accuracy ($\%$) of students (WRN16-1) for knowledge distillation methods, trained with SP and a teacher (WRN16-3) on CIFAR-10. ``T'' and ``S'' denote teacher and student, respectively. ``g'' and ``l'' denote using global and local feature distillation, respectively. ``Student'' is a result of WRN16-1 trained from scratch.}
\label{figure:SP_results}
\end{figure}

\begin{table}[htb!]
\centering
\caption{ECE ($\%$) and NLL ($\%$) for various knowledge distillation methods with SP on CIFAR-10. ``$\mathrm{g}$'' and ``$\mathrm{l}$'' denote using global and local feature distillation, respectively. The results (ECE, NLL) for WRN16-3 and WRN28-1 teachers are (1.469$\%$, 44.42$\%$) and (2.108$\%$, 64.38$\%$), respectively.}

\begin{center}
\scalebox{0.73}{
\begin{tabular}{c |c |c c | c c}

\hline
\centering

\multirow{2}{*}{Setup} & \multirow{2}{*}{Method} & \multicolumn{2}{c|}{w/o SP}&  \multicolumn{2}{c}{w/ SP}\\ 
& & ECE & NLL & ECE & NLL \\
\hline

\multirow{4}{*}{(a)} &AT \cite{zagoruyko2016paying}& 1.978 & 60.48 & 1.861 (\textcolor{blue}{-0.118}) & 56.22 (\textcolor{blue}{-4.26}) \\

&AFD \cite{ji2021show}& \textbf{1.890} & \textbf{56.71} & 1.881 (\textcolor{blue}{-0.010}) & 56.73 (\textcolor{blue}{-0.02})  \\

&AMD (g) & 1.933 & 59.67 & 1.808 (\textcolor{blue}{-0.125}) & 54.74 (\textcolor{blue}{-4.93}) \\

&AMD (g+l) & 1.895 & 57.60 & \textbf{1.803} (\textcolor{blue}{-0.092}) & \textbf{53.80} (\textcolor{blue}{-3.80}) \\

\hline
 
\multirow{5}{*}{(b)} &AT \cite{zagoruyko2016paying}& 2.156 & 67.14 & 2.095 (\textcolor{blue}{-0.060}) & 65.38 (\textcolor{blue}{-1.75}) \\

&AFDS \cite{wang2019pay}& 2.197 & 68.53 & 2.128 (\textcolor{blue}{-0.069}) & 66.61 (\textcolor{blue}{-1.92})  \\

&AFD \cite{ji2021show}& 2.143 & 66.05 & 2.118 (\textcolor{blue}{-0.024}) & 65.39 (\textcolor{blue}{-0.66})  \\

&AMD (g) & \textbf{2.117} & \textbf{66.47} & 2.058 (\textcolor{blue}{-0.059}) & 63.37 (\textcolor{blue}{-3.10}) \\

&AMD (g+l) & 2.123 & 67.51 & \textbf{2.043} (\textcolor{blue}{-0.080}) & \textbf{63.23} (\textcolor{blue}{-4.28}) \\

\hline

\end{tabular} }
\end{center}

\label{table:addSP_result}
\end{table}

\begin{figure}[ht!] 
\includegraphics[scale=0.31] {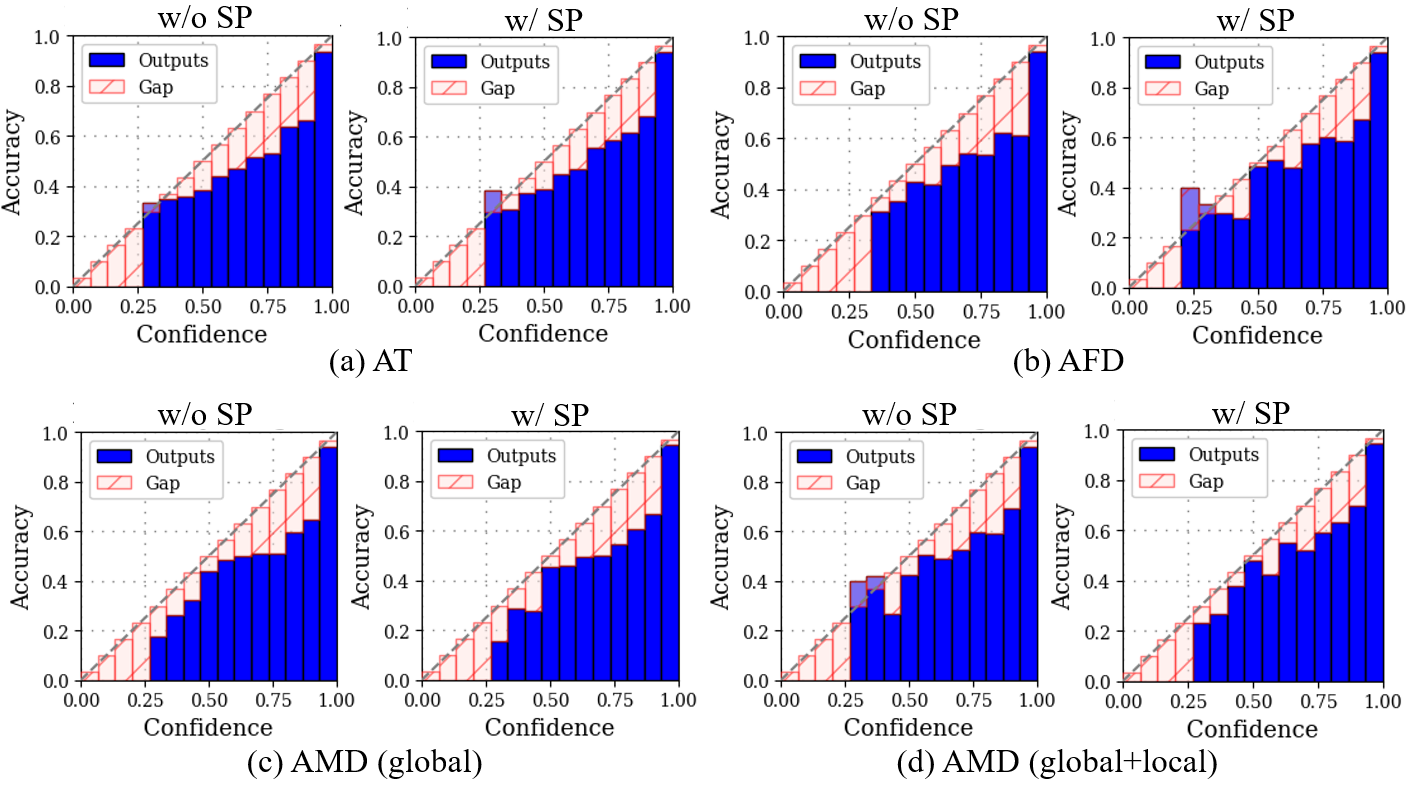}
\centering
\caption{Reliability diagrams of students (WRN16-1) for knowledge distillation methods, trained with SP and a teacher (WRN16-3) on CIFAR-10. For the results of each method, the left is the result without SP, and the right is with SP.}
\label{figure:sp_reliability}
\end{figure}


To demonstrate how AMD can perform with the other distillation methods, we adopt SP \cite{tung2019similarity} which is not an attention based distillation method. A teacher is trained with the original training set and learns from scratch. SP \cite{tung2019similarity} is applied while a student is being trained. We compare with baselines, depicted in Figure \ref{figure:SP_results}. In all cases, with SP, the accuracy is increased. Compared to the other attention based methods, AMD gets more gains by SP. Therefore, AMD can be enhanced and can perform well with the other distillation methods such as SP. We additionally analyzed the reliability described in Table \ref{table:addSP_result}. AMD (global+local) with SP shows the lowest ECE and NLL values. It verifies that AMD with SP can generate a model having higher reliability with better accuracy. Thus, the proposed method can be used with an additional distillation method. Also, the proposed method with SP can perform with different combinations of teacher and student with well-calibrated results. As illustrated in Figure \ref{figure:sp_reliability}, with SP \cite{tung2019similarity}, AT \cite{zagoruyko2016paying} and AFD \cite{ji2021show} produce more overconfident predictions, compared to AMD (global+local) with SP \cite{tung2019similarity} that gives the best calibration performance. Conclusively, our empirical findings reveal that AMD can perform with other distillation methods such as SP \cite{tung2019similarity} to generate more informative features for distillation from teacher to student.

 
\section{Conclusion} \label{sec:conclusion}
In this paper, we proposed a new type of distillation loss function, AMD loss, which uses the angular distribution of features. We validated the effectiveness of distillation with this loss, under the setting of multiple teacher-student architecture combinations of KD in image classification. Furthermore, we have confirmed that the proposed method can be combined with previous methods such as fine-grained feature, various augmentation methods, and other types of distillation methods.

In future work, we aim to extend the proposed method to explore the distillation effects with different hypersphere feature embedding methods \cite{wang2018cosface, deng2019arcface}. Also, we plan to extend AMD to different approaches in image classification, such as vision transformer \cite{dosovitskiy2020image} and
MLP-mixer \cite{tolstikhin2021mlp} that are not based on convolutional neural network. In addition, our approach could provide insights for further advancement in other applications such as object detection and semantic segmentation. 

\section*{Acknowledgements}
This material is based upon work supported by the Defense Advanced Research Projects Agency (DARPA) under Agreement No. HR00112290073. Approved for public release; distribution is unlimited.









%
{
\bibliographystyle{elsarticle-num}
\bibliography{manuscript}
}

\end{document}